\documentclass[fleqn,10pt]{wlscirep}
\usepackage[utf8]{inputenc}
\usepackage[T1]{fontenc}
\usepackage{multirow}

\usepackage{subfigure}
\usepackage{enumitem}
\setlist{nolistsep}
\setlist{nosep}
\usepackage{colortbl}

\title{PND-Net: Plant Nutrition Deficiency and Disease Classification using Graph Convolutional Network}

\author[1,*]{Asish Bera}
\author[2,3]{Debotosh Bhattacharjee}
\author[3,4,5]{Ondrej Krejcar}
\affil[1]{Department of Computer Science and Information Systems, BITS Pilani, Pilani Campus, Rajasthan, 333031, India}
\affil[2]{Department of Computer Science and Engineering, Jadavpur University, Kolkata, 700032, WB}

\affil[3]{Faculty of Informatics and Management University of Hradec Kralove, Hradec Kralove, Czech Republic}
\affil[4]{
Skoda Auto University, Na Karmeli 1457, 293 01 Mlada Boleslav, Czech Republic}
\affil[5]{Malaysia Japan International Institute of Technology (MJIIT), Universiti Teknologi Malaysia, Kuala Lumpur, Malaysia}

\affil[1, *]{asish.bera@pilani.bits-pilani.ac.in}
\affil[2]{debotosh.bhattacharjee@jadavpuruniversity.in}
\affil[3]{ondrej.krejcar@uhk.cz}
\begin{abstract}
Crop yield production could be enhanced for agricultural growth if various plant  nutrition deficiencies, and diseases are identified and detected at early stages. Hence, continuous health monitoring of plant is very crucial for handling plant stress. The deep learning methods have proven its superior performances in the automated detection of plant diseases and nutrition deficiencies from visual symptoms in leaves. This article proposes a new deep learning method for plant nutrition deficiencies and disease classification using a graph convolutional network (GNN), added upon a base  convolutional neural network (CNN). Sometimes, a global feature descriptor might fail to capture the vital region of a diseased leaf, which causes inaccurate classification of disease. To address this issue,  regional feature learning is crucial for a holistic feature aggregation. In this work, region-based feature summarization at multi-scales is explored using  spatial pyramidal pooling for discriminative feature representation. 

Furthermore, a GCN is developed to capacitate learning of finer details for classifying plant diseases and insufficiency of nutrients. The proposed method, called \textbf{P}lant \textbf{N}utrition Deficiency and \textbf{D}isease \textbf{Net}work (PND-Net), has been  evaluated on two public datasets for nutrition deficiency, and two  for disease classification using four backbone CNNs. The best classification performances  of the proposed PND-Net are as follows: (a) 90.00\% Banana and 90.54\% Coffee nutrition deficiency; and (b) 96.18\% Potato diseases and 84.30\% on PlantDoc datasets using Xception backbone. Furthermore,  additional experiments have been carried out  for generalization, and the proposed method has achieved state-of-the-art performances on two public datasets, namely the Breast Cancer Histopathology Image Classification (BreakHis 40X: 95.50\%, and BreakHis 100X: 96.79\% accuracy) and Single cells in Pap smear images for cervical cancer classification (SIPaKMeD: 99.18\% accuracy). Also, the proposed method has been evaluated using five-fold cross validation and achieved improved performances on  these datasets. Clearly, the proposed PND-Net  effectively boosts the performances of automated health analysis of various plants in real and intricate field environments, implying  PND-Net's aptness for agricultural growth as well as human cancer classification.

\end{abstract}

\begin{document}

\flushbottom
\maketitle
\thispagestyle{empty}

%\noindent Please note: Abbreviations should be introduced at the first mention in the main text – no abbreviations lists. Suggested structure of main text (not enforced) is provided below.
\noindent \textbf{Keywords:} {Agriculture, Convolutional Neural Network, Graph Convolutional Network, Plant Disease, Nutrition Deficiency,  Cancer Classification, Spatial Pyramid Pooling. }

\section{Introduction}

Agricultural production plays a crucial role  in the  sustainable economic and societal growth of a country. High-quality crop yield production is essential for satisfying global food demands and better health. However, several key factors, such as  environmental barriers, pollution,  and  climate change,  adversely affect crop yield and quality. Nevertheless,  poor soil-nutrition management causes severe plant stress, leading to different diseases and resulting in a substantial financial loss. Thus, plant nutrition diagnosis and disease detection at an early stage is of utmost importance for overall health monitoring of plants \cite{jung2023construction}. 
Nutrition management in agriculture is a decisive task for maintaining the  growth of plants. In recent times, it has been witnessed the success of  machine learning (ML) techniques for developing decision support systems over traditional manual supervision of agricultural yield. Moreover, nutrient management is critical  for improving production growth, focusing on a robust and low-cost solution. Intelligent automated systems based on ML  effectively build more accurate predictive models, which are relevant for improving agricultural production. 

Nutrient deficiency in plants exhibits certain visual symptoms and may cause of poor crop yields \cite{aiswarya2023plant}.  Diagnosis of plant nutrient inadequacy using deep learning and related intelligent methods is an emerging area in precision agriculture and plant pathology \cite{yan2022nutrient}. Automated detection and classification of nutrient deficiencies using computer vision and artificial intelligence  have been studied in the recent literature \cite{sudhakar2023computer}, \cite{noon2020use}, \cite{waheed2022deep}, \cite{barbedo2019detection}, \cite{shadrach2023optimal}. 
Diagnosis  of nutrient deficiencies in various plants (e.g., rice, banana, guava, palm oil, apple, lettuce, etc.) is  vital, because soil ingredients often can not provide the nutrients as required for the growth of plants 
\cite{sathyavani2022classification}, \cite{haris2023nutrient}, \cite{munir2023application}, \cite{lu2023lettuce}. Also, early stage detection of leaf diseases (e.g., potato, rice, cucumber, etc.) and pests are essential to monitor crop yield production \cite{omer2023lightweight}.
A few approaches on disease detection and nutrient deficiencies in rice leaves have been developed and studied in recent times \cite{kumar2023automated},  \cite{senjaliya2023comparative}, \cite{ennaji2023machine}, \cite{ennaji2023machine}, \cite{rathnayake2023green},\cite{asaari2023detection}.
Hence, monitoring  plant health, disease, and nutrition inadequacy could be  a  challenging  image classification problem in artificial intelligence (AI) and machine learning (ML)\cite{tavanapong2022artificial}. 

This paper proposes a deep learning method for plant health diagnosis by integrating a graph convolutional network (GCN)  upon a backbone deep convolutional neural network (CNN). The complementary discriminatory features of different local regions of input leaf images are aggregated into a holistic representation for plant nutrition and disease classification. 
The GCNs were originally developed for semi-supervised node classification \cite{kipf2016semi}. Over time, several variations of GCNs have been developed for graph structured data \cite{zhang2019graph}.
Furthermore, GCN is effective for message propagation for image and video data in various applications. In this direction, several works have been developed for image recognition using GCN \cite{bera2022sr}, \cite{qu2023graph}. However,  little research attention has been given to adopting GCN especially for plant disease prediction and nutrition monitoring \cite{khlifi2023graph}.  Thus, in this work,  we have studied the effectiveness of GCN in solving the current problem of plant health analysis regarding  nutrition deficiency and disease classification of several categories of plants. 

The proposed method, called Plant Nutrition Deficiency and Disease Network (PND-Net),  attempts to establish a correlation between different regions  of the leaves  for identifying  infected and defective regions at multiple granularities. For this intent, region pooling in local contexts and spatial pooling in a pyramidal structure,  have been explored for a holistic feature representation of subtle discrimination of plant health conditions. Other existing approaches have built the graph-based correlation directly upon the CNN features, but they have often failed to capture finer descriptions of the input data. In this work, we have integrated two different feature pooling techniques for generating node features of the graph. As a result, this mixing enables an enhanced feature representation which is further improved by graph layer activations in the hidden layers in the GCN. The effectiveness of the proposed strategy has been analysed with rigorous experiments on two plant nutrition deficiency and two plant disease classification datasets. In addition, the method has been tested on two different human cancer classification tasks for the generalization of the method. The key contributions of this work are:
\begin{itemize}  
 \item A deep learning method, called PND-Net, is devised by integrating a graph convolutional module upon a  base CNN  to enhance the feature representation for improving the classification performances of  unhealthy leaves. 

  \item A combination of fixed-size region-based  pooling with multi-scale spatial pyramid pooling progressively enhances the feature aggregation for building a spatial relation between the regions via the neighborhood nodes of a spatial graph structure. %

\item Experimental studies have been  carried out for validating the proposed method on four public datasets,  which have been tested for plant disease classification, and nutrition deficiency classification. For generalization of the proposed method, a few experiments have been conducted on the cervical cancer cell (SIPaKMeD) and breast cancer histopathology image (BreakHis 40X and 100X) datasets. The  proposed PND-Net  has achieved state-of-the-art performances on these six public datasets of different categories. 

 \end{itemize}
 
The rest of this paper is organized as follows: Section \ref{rel_work} summarizes related works.  Section \ref{proposed} describes the proposed methodology. The experimental results are showcased in Section \ref{experiments}, followed by the conclusion in Section \ref{conclusion}.

\section{Related Works} \label{rel_work} 
Several works have been contributed to plant disease detection, most of which were  tested on controlled datasets, acquired in a laboratory set-up. 
Only a few works have developed unconstrained datasets considering realistic field conditions, which have been studied in this work. Here, a precise study of recent works has been briefed.

\subsection{Methods on Plant Nutrition Deficiencies}
Bananas are one of the widely consumed staple foods  across the world. An image dataset depicting the visual deficiency symptoms of eight essential nutrients, namely,  boron, calcium, iron, potassium, manganese, magnesium, sulphur and zinc has been developed  \cite{sunitha2023fully}. This dataset has been tested in this proposed work. 
The CoLeaf dataset contains images of coffee plant leaves and is tested for nutritional deficiencies recognition and classification \cite{tuesta2023coleaf}.
 The nutritional status of oil palm leaves, particularly  the status of chlorophyll and macro-nutrients (e.g., N, K, Ca, and Mg) in the leaves from proximal multi spectral images, have been evaluated using machine learning techniques \cite{chungcharoen2022machine}. The identification and categorization of common macro-nutrient  (e.g., nitrogen, phosphorus,  potassium, etc.) deficiencies in rice plants has been addressed  \cite{rathnayake2023green}, \cite{bhavya2023fertilizer}.
The percentage of micro nutrients deficiencies in rice plants using CNNs and Random Forest (RF) has been estimated  \cite{bhavya2023fertilizer}. 
Detection of biotic stressed rice leaves and  abiotic stressed leaves caused by NPK (Nitrogen, Phosphorus, and Potassium) deficiencies have been experimented with using CNN \cite{dey2022comparative}. 

A supervised  monitoring system of  tomato leaves for predicting nutrient deficiencies using a CNN for recognizing and to classify the type of nutrient deficiency in tomato plants and achieved 86.57\% accuracy \cite{cevallos2020vision}. 
The nutrient deficiency symptoms have been  recognized in RGB images by using CNN-based (e.g., EfficientNet) transfer learning on  orange with 98.52\% accuracy and sugar beet with 98.65\% accuracy  \cite{espejo2022using}. 
Nutrient deficiencies in rice plants have reported 97.0\% accuracy by combining CNN and reinforcement learning  \cite{wang2021classification}. 
The  R-CNN object detector has achieved accuracy of 82.61\%  for identifying nutrient deficiencies in chili leaves \cite{bahtiar2020deep}.
Feature aggregation schemes by combining the features with HSV and RGB for color, GLCM and LBP for texture, and Hu moments and centroid distance for shapes have been examined for nutrient deficiency identification in chili plants \cite{rahadiyan2023feature}. However, this method performed the best using a CNN with 97.76\% accuracy. An ensemble of CNNs has reported  98.46\% accuracy  for detecting groundnut plant leaf images \cite{aishwarya2023ensemble}.
An intelligent robotic system  with a wireless control to monitor the nutrition essentials  of spinach plants in the greenhouse has been evaluated with 86\% precision \cite{nadafzadeh2024design}.
The nutrient status and health conditions of the  Romaine Lettuce plants in a hydroponic setup using a CNN have been tested with 90\% accuracy \cite{desiderio2022health}.
The identification and categorization of common macro-nutrient (e.g., nitrogen, phosphorus,  potassium, etc.) deficiencies in rice plants using  pixel ratio analysis in HSV color space has been evaluated with more than 90\% accuracy \cite{rathnayake2023green}.
A method for estimating leaf nutrient concentrations of citrus trees  using unmanned aerial vehicle (UAV) multi-spectral images has been developed and tested by a gradient-boosting regression tree model with moderate precision \cite{costa2022determining}. 

 \subsection{Approaches on Plant Diseases}

 The classification of  healthy and diseased citrus leaf images using a (CNN) on the Platform as a Service (PaaS) cloud has been developed. The method has been tested using pre-trained backbones and proposed CNN, and attained 98.0\% accuracy and   99.0\% F1-score \cite{lanjewar2023cnn}. A modified transfer learning (TL) method using three pre-trained CNN has been tested for potato leaf disease detection and the DensNet169 has achieved 99.0\% accuracy \cite{lanjewar2023modified}.  
Likewise, a CNN-based transfer learning method has been adapted for detecting 
powdery mildew disease with 98.0\% accuracy in bell pepper leaves \cite{dissanayake2023detection},  and woody fruit leaves with 85.90\% accuracy \cite{wu2022research}. A two-stage transfer learning method has combined Faster-RCNN for leaf detection and CNN for  maize plant disease recognition in a natural environment and obtained  99.70\% F1-score \cite{liu2022research}. 
A  hybrid model integrating a CNN and random forest (RF) for multi-classifying rice hispa disease into distinct  intensity levels  \cite{kukreja2023deepleaf}. A method of multi-classification of rice hispa illness has attained accuracy of
97.46\% using  CNN and RF \cite{kukreja2023deepleaf}. An improved YOLOv5 network has been developed for cucumber leaf diseases and pest detection  and reported 73.8\% precision \cite{omer2023lightweight}.
A fusion of VGG16 and AlexNet architecture has attained 95.82\% testing accuracy  for pepper leaf  disease classification \cite{bezabih2023cpd}. Likewise, the disease classification of black pepper has gained 99.67\% accuracy  using ResNet-18 \cite{kini2024early}. A ConvNeXt with an attention module, namely CBAM-ConvNeXt has improved the performance with 85.42\% accuracy for classifying soybean leaf disease   \cite{wu2023classification}.
A channel extension residual structure with an adaptive channel attention mechanism and a bidirectional information fusion block for leaf disease classification \cite{ma2024ercp}. This technique has brought off  99.82\% accuracy on the plantvillage dataset.
A smartphone application has been developed for detecting habanero plant disease and obtained 98.79\% accuracy \cite{babatunde2024novel}. In addition, an ensemble  method for crop monitoring system  to identify plant diseases at the early stages using IoT enabled system has been presented with the best precision of 84.6\%  \cite{nagasubramanian2021ensemble}.
 A dataset comprising five types of disorders of apple orchards has been developed, and the best accuracy is 97.3\%, which has been tested using CNN \cite{nachtigall2016classification}.
A lightweight model using ViT structure has been developed for rice leaf disease classification and attained 91.70\% F1-score  \cite{borhani2022deep}.

\subsection{ Methods on Graph Convolutional Networks (GCN) } 
Though several deep learning approaches have been developed for plant health analysis yet, little progress has been achieved using GCN for visual recognition of plant diseases \cite{aishwarya2023dataset}.
 The SR-GNN integrates relation-aware feature representation leveraging context-aware attention with the GCN module \cite{bera2022sr}. Cervical cell classification methods have been developed by exploring the potential correlations of clusters through GCN \cite{shi2021cervical} and  feature rank analysis \cite{fahad2024enhancing}. On the other side, fusion of multiple CNNs, transfer learning and other deep learning methods have been developed for early detection of breast cancer  \cite{lanjewar2024fusion}. This fusion method has achieved F1 score of 99.0\% on ultrasound breast cancer dataset using VGG-16. 
In this work,  a GCN-based  method has been developed by capturing the regional importance of local contextual features in solving plant disease recognition and human cancer image classification challenges.  

\section{{Proposed Method}} \label{proposed}
\begin{figure}
\centering
\includegraphics[width= 0.98 \textwidth ]{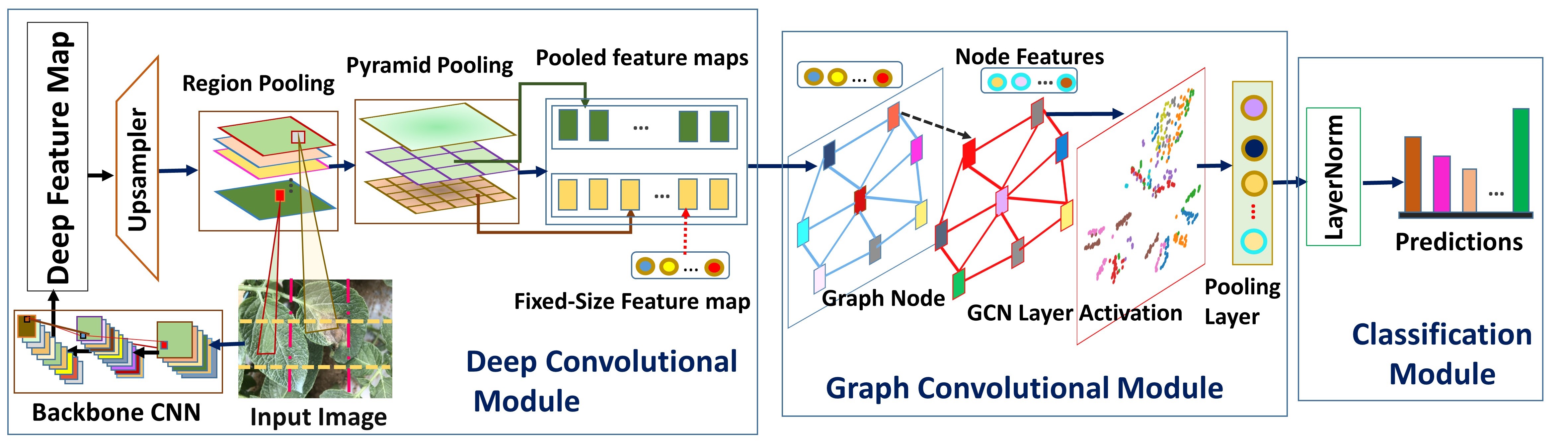}
\caption{Proposed GCN-based method, PND-Net for visual classification of plant disease and nutrition inadequacy.} \label{model}
\end{figure}

The proposed method, called PND-Net, combines deep features using CNN and GCN in an end-to-end pipeline as shown in Fig. \ref{model}. Firstly, a backbone CNN computes high-level deep features from input images. Then, a GCN is included upon the  CNN for refining deep features using  region-based pooling and pyramid pooling strategies for capturing finer details of contextual regions at multiple scales.  Finally, a precise  feature map is built for improving the  performance. 

\subsection{Background of Graph Convolutional Network (GCN)}
GCNs have  widely been used for several domains and  applications such as node classification, edge attribute modeling, citation networks, knowledge graphs, and several other tasks through graph-based representation. 
A GCN could be formulated by stacking multiple graph convolutional layers with non-linearity upon traditional convolutional layers, i.e., CNN. In practice, this kind of stacking of GCN layers at a deeper level of a network enhances the model's  learning capacity. Moreover,  graph convolutional layers are effective for alleviating overfitting issue and can  address the vanishing gradient problem by adopting the normalization trick, which is a foundation of modeling GCN. A widely used multi-layer GCN  algorithm was proposed by Kiff and Welling \cite{kipf2016semi}, which has been adopted here. It explores an efficient  and fast layer-wise propagation method relying on the first-order approximation of spectral convolutions on graph structures. It is  scalable and apposite for semi-supervised node classification from graph-based data. 
A linear formulation of a GCN could be simplified which, in turn, is capable of parameter optimization at each layer by convolution with filter $g_{\theta}$ and $\theta$ parameters, which can further be optimized with a single parameter. Here, a simplified graph convolution has been concisely defined   \cite{kipf2016semi}.

\begin{equation} \label{eqn0}  %
\centering
g_{\theta} \ast X\approx \theta \big(I_P+D^{-0.5}AD^{-0.5} \big)X
\end{equation}

\noindent The graph Laplacian ($\Psi$) could further be normalized to mitigate the vanishing gradients within a network. 
\begin{equation} \label{eqn1}  %
\centering
\Psi= I_P+D^{-0.5}AD^{-0.5} \rightarrow \tilde{\mathbf{D}}^{-0.5} \tilde{\mathbf{A}} \tilde{\mathbf{D}}^{-0.5} 
\end{equation}
\noindent where the binary adjacency matrix $\tilde{\mathbf{A}}=\mathbf{A}+\mathbf{I}_{{P}}$ denotes $\mathbf{A}$ with self-connections and $\mathbf{I}_{{P}}$ is the identity matrix, and degree matrix is $\tilde{\mathbf{D}}_{ii}= \sum_{j}^{\hspace{0.2 cm}}  \tilde{\mathbf{A}}_{ij}$, and $X$ is an input data/signal to the graph.
The simplified convoluted signal matrix $\Omega$ is given as
\begin{equation} \label{eqn2}  %
\centering
\Omega= \tilde{\mathbf{D}}^{-0.5} \tilde{\mathbf{A}} \tilde{\mathbf{D}}^{-0.5} X \Theta
\end{equation}
where input features $X \in \mathbb{R}^{{P}\times{C}}$, filter parameters $\Theta \in \mathbb{R}^{{C}\times{F}}$, and $\Omega \in  \mathbb{R}^{{P}\times{F}}$ is the convoluted signal matrix. Here, P is the number of nodes, C is the input channels, F is the filters/feature maps. Now, this form of graph convolution (eqn \ref{eqn2}) is applied to address the current problem and is described in Section \ref{GCN1}. 

\subsection{Convolutional Feature Representation}

A standard backbone CNN is used for deep feature extraction from an input leaf image, denoted with the class label $I_l$ $\in$ $\mathbb{R}^{h\times w\times 3}$ is passed through a base CNN for extracting the feature map, denoted as $\mathbf{F}$ $\in$ $\mathbb{R}^{h\times w\times C}$ where $h$, $w$, and $C$ imply the height, width, and channels, respectively. 
However, the squeezed high-level feature map is not suitable for describing local non-overlapping regions. Hence, the output base feature map is spatially up-sampled to $\mathbf{F}$ $\in$ $\mathbb{R}^{H\times W\times C}$ and $\omega$ number of distinct small regions are computed, given as $\mathbf{F}$ $\in$ $\mathbb{R}^{\omega \times h\times w\times C}$. These regions represents complementary information at different spatial contexts. However, due to fixed dimensions of  regions, the importance of each region is uniformly distributed, which could be tuned further for extracting more distinguishable information.  A simple pooling technique could further be applied  at multiple scales for enhancing the spatial feature representation.  
 For this intent, the region-pooled feature vectors are reshaped to convert them into an aggregated spatial feature space upon which multi-scale pyramidal pooling is possible. In addition, this kind of feature representation captures overall spatiality to understand the informative features holistically and solve the current problem. 

\subsubsection{Spatial Pyramid Pooling (SPP)}
The SPP layer was originally introduced to alleviate the fixed-length input constraints of conventional deep networks, which effectively boosted the model's performance  \cite{he2015spatial}. 
Generally,  a SPP layer is added upon the last convolutional layer of a backbone CNN. This pooling layer generates a fixed-length feature vector and afterward passes the feature map to a fully connected or classification layer. The SPP  enhances  feature aggregation capability at a deeper layer of a network. Most importantly, SPP applies multi-level spatial bins for  pooling while preserving the spatial relevance of the feature map. It provides a robust solution through performance enhancement of diverse computer vision problems, including plant/leaf image recognition.

A typical region polling technique loses its spatial information while passing though a global average pooling (GAP) layer for making compatible with  and plugging in the GCN. As a result, a region pooling with a GAP layer aggressively eliminates informativeness of regions and their correlation, and thus often it fails to build an effective feature vector. Also, the inter-region interactions are ignored with a GAP layer upon only region-based pooling. Therefore, it is essential to  correlate  the inter-region interactions for selecting essential features, which could further be enriched and propagated through the GCN layer activations.  

Our objective is to utilize the advantage of multi-level pooling  at different pyramid levels of $n \times n$ bins on the top of fixed-size regions of the input image. As a result, the spatial relationships between different image regions are preserved,  thereby escalating the learning capacity of the proposed PND-Net. The input feature space prior to pyramid pooling is  given as $\mathbf{F}^{\omega \times (HW)\times C}$, which has been derived from $\mathbf{F}^{\omega \times H\times W\times C}$. It enables the selection of contextual features of neighboring regions (i.e., inter-regions) through pyramid pooling simultaneously.
This little  adjustment in the spatial dimension  of input features prior to pooling captures the interactions between the local regions of input leaf disease. Experimental results reflect that pyramidal pooling indeed elevates image classification accuracy gain over region pooling only. 
\begin{equation} \label{eqSPP}
\centering
\vspace{- 0.5 cm}
{F}_{SPP}=\textit{PyramidPooling}\Big(\textbf{F}_{\delta_i\times \delta_i}; \textbf{F}_{\delta_j\times \delta_j}\Bigl)
\end{equation}
\noindent where $\delta_i$ and $\delta_j$ define the window sizes, which enable to pool a total of $P=(i\times i) + (j\times j)$ feature maps after SPP, given as $\mathbf{F}^{P\times C}$. These feature maps are further fed into a GCN module, described next.  The key components of  proposed method are pictorially ideated in Fig. \ref{model}.

\subsection{Graph Convolutional Network (GCN) } \label{GCN1}

A graph $G=({P},E)$, with  ${P}$ nodes and   $E$ edges, is constructed for feature propagation. A GCN is applied for building a spatial relation between the features through graph $G$. The nodes are characterized by deep feature maps, and the  output $\mathbf{C}$ with the convoluted features per node. The  edges $E$ are described by an un-directed adjacency matrix $\mathbf{A} \in \mathbb{R}^{{P}\times{P}}$ for representing node-level interactions. This graph convolution has been  applied to  $F_{SPP}$ (i.e., $\mathbf{F}^{P\times C}$), described above. 
 The layer-wise feature propagation rule is defined as: 
\begin{equation} \label{eq5}  %
\centering
 \mathbf{G}^{(l+1)} = \sigma\left(\hat{\tilde{\mathbf{A}}}\mathbf{G}^{(l)}\mathbf{W}^{(l)}\right); \hspace{0.2 cm}  \text{with}  \hspace{0.2 cm}  \mathbf{G}^{(0)}=\mathbf{F}^{P\times C}, \hspace{0.2 cm}   \text{and} \hspace{0.2 cm} \mathbf{G}^{(L)}=\mathbf{F}^{P\times C}
\end{equation}
\noindent $l=0, 1, \dots, L-1$ is the number of layers, $\mathbf{W}^{(l)}$ is a weight matrix for the $l$-th layer.  A non-linear activation function (\textit{e.g.}, ReLU) is denoted by $\sigma(.)$. The symmetrically normalized adjacency matrix is
$\hat{\tilde{\mathbf{A}}}=Q\tilde{\mathbf{A}}Q; $ and $Q=\tilde{\mathbf{D}}^{-1/2}$ is the diagonal node degree matrix of $\tilde{\mathbf{A}}$ (defined in eqn. \ref{eqn2}). 
Next, the reshaped convolutional feature map $\textbf{F}$ is fed into  two layers of graph convolutions,  subsequently which is capable of capturing local neighborhoods via the non-linear activations of rectified linear unit  (ReLU) in the graph convolutional layers. The  dimension of the output feature maps remains the same input of GCN layers, i.e., $ \mathbf{G}^{(L)}\rightarrow \textbf{F}_{G}$ $\in \mathbb{R}^{{P}\times{C}}$. However, the node features could be squeezed to a lower dimension, which may lose essential information pertinent to spatial modeling. Hence, the channel dimension is kept uniform within the network pipeline in our study. Afterward, the graph-based transformed  feature maps ($\textbf{F}_{G}$)  are pooled using a GAP for selecting the most discriminative channel-wise feature maps of the nodes.

\begin{figure}
\centering
{
    \includegraphics[width=0.32\textwidth, height=3.5 cm]{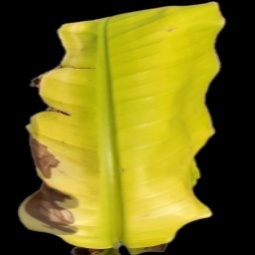}    \hfill
    \includegraphics[width=0.32\textwidth, height=3.5 cm]{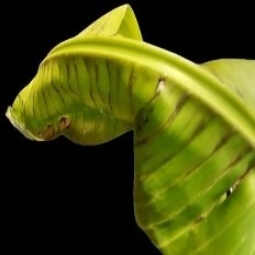}   \hfill  
    \includegraphics[width=0.32\textwidth, height=3.5 cm]{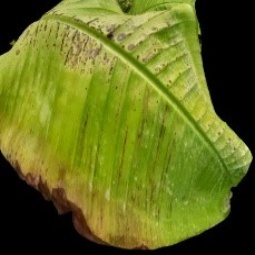}  }
\caption{Sample images of banana dataset showing the nutrition deficiency of iron, calcium, and magnesium.}
    \label{banana}
\end{figure}
\begin{figure}
\centering
{
\includegraphics[width=0.32\textwidth, height= 3.5 cm]{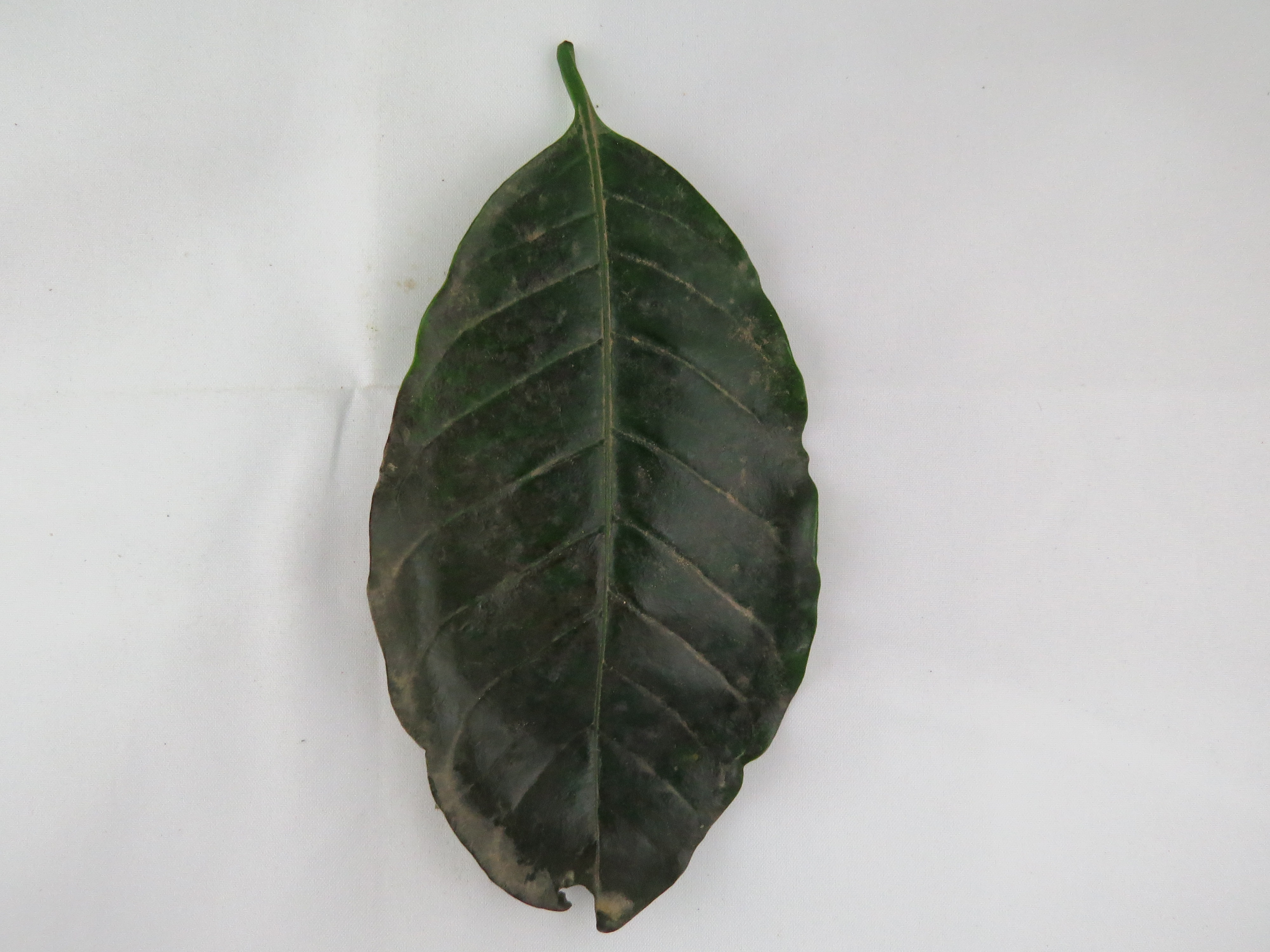}    \hfill
    \includegraphics[width=0.32\textwidth, height= 3.5 cm]{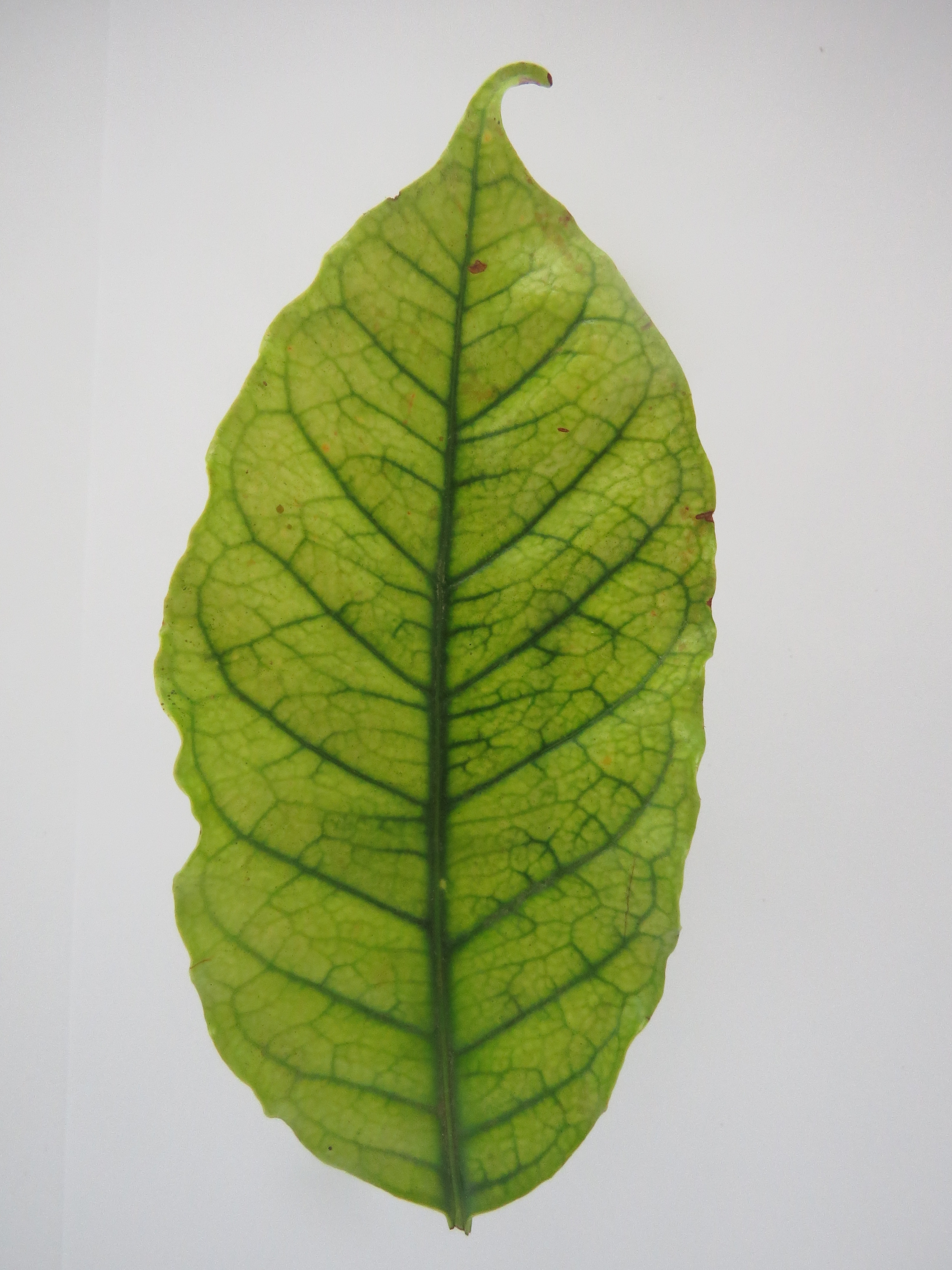}   \hfill  
   \includegraphics[width=0.32\textwidth, height= 3.5 cm]{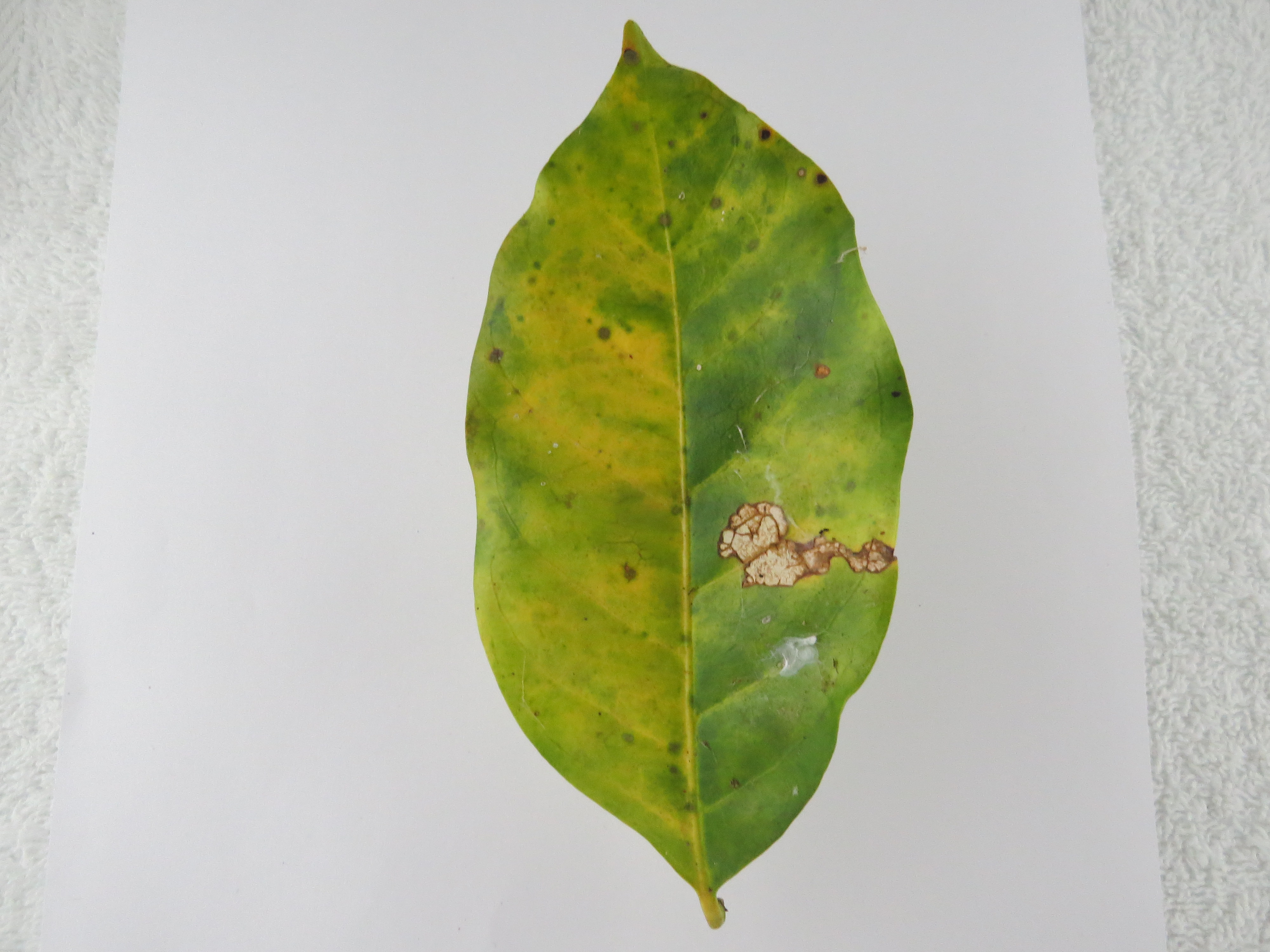}  }
\caption{Sample images of Coffee nutrition deficiency of boron, manganese, and nitrogen.}
    \label{coffee}
    \vspace{ -0.2 cm}
\end{figure}

\subsection{Classification Module} \label{class1}

Generally, regularization is a standard way to tackle the training-related challenges of any network, such as overfitting. Here, the layer normalization and dropout layers are interposed for handling overfitting issues as a regularization technique.  Lastly,  ${F}_{final}$ is passes through a \textit{softmax} layer for computing the output probability of the predicted class-label $\bar{b}$,  corresponding to the actual-label $b \in Y$ of object classes $Y$. 

\begin{equation} \label{eq3}
\centering
\textit{F}_{final}=\textit{Regularization} \Bigl( GAP\big({F}_{G}\big) \Bigl)
,  \hspace{0.3 cm} and \hspace{0.3 cm}  {Y}_{pred}={Softmax}\Bigl(\textit{F}_{final}\Bigl). 
\end{equation}

The categorical cross-entropy loss function ($\mathcal{L}_{CE}$) and the Stochastic Gradient Descent (SGD) optimizer  with $10^{-3}$ learning rate has been chosen for experiments.
\vspace{- 0.2 cm}
 {
\begin{equation} \label{CELoss} 
\begin{split}
\mathcal{L}_{CE}= - \sum_{i=1}^{N} Y_{i}.log\hat{Y}_i
\end{split}
%\vspace{ -0.5cm}
\end{equation}
where $Y_i$ is the actual class label and $log\hat{Y}_i$ is the predicted class label by using softmax activation function $\sigma(.)$ in the classification layer, and $N$ is the total number of classes.
}

\section{Results and Performance Analysis} \label{experiments}

At first, the implementation description is provided,  followed by a summary of  datasets. The experiments have been conducted using conventional classification and cross validation methods. The performances are evaluated using the standard well-known metrics: accuracy, precision, recall, and F1-score (eq. \ref{metric}). 

 {
\begin{equation} \label{metric}
   \begin{split}
{Accuracy}=\frac{TP+TN}{TP+TN+FP+FN}\ \\  
 {Precision }=\frac{TP}{TP+FP}\  \\
 {Recall} = \frac{TP}{TP+FN}\  \\
 {F1}\text{-}{Score }= 2\times\frac{{Precision}\times {Recall}}{{Precision} + {Recall}}  
 \end{split} 
 \end{equation}
  }
{\noindent where TP is the number of true positive, TN is the number of true negative, FP is the number of false positive, and FN is
the number of false negative. However, accuracy is not a good assessment metric when the data distributions among the classes are imbalanced. To overcome such misleading evaluation, the precision and recall are useful metrics, based on which F1-score is measured. These three metrics are widely used for evaluating the predictive performance when classes are imbalanced.  In addition, we have evaluated the performance using confusion matrix which provides a reliable performance assessment of our model. The performances have been compared with existing methods, discussed below. }
{
\subsection {Implementation Details} \label{implemnt}
A concise description about the model development regarding the hardware resources, software implementation  data distribution, evaluation protocols, and related details are furnished below for easier understanding.
}

\vspace{- 0.3 cm}

\subsubsection{Summary of Convolutional Network Architectures}
The Inception-V3, Xception, ResNet-50, and MobileNet-V2  backbone CNNs with pre-trained ImageNet weights are used for convolutional feature computation from the input images. The Inception module focuses on increasing network depth using 5$\times$5, 3$\times$3, and 1$\times$1 convolutions \cite{szegedy2015going}. Again,  5$\times$5 convolution has been replaced by factorizing into 3$\times$3 filter sizes \cite{szegedy2016rethinking}. Afterward, the Inception module is further decoupled the channel-wise and spatial correlations by point-wise and depth-wise separable convolutions, which are the building block of Xception architecture \cite{chollet2017xception}. The separable convolution follows the depth-wise convolution for spatial (3$\times$3 filters) and point-wise convolution (1$\times$1 filters) for cross-channel aggregation into a single feature map. The Xception  is a three-fold architecture  developed with depth-wise separable convolution layers with residual connections. Whereas, the residual connection a.k.a. shortcut connection is the central idea of deep residual learning framework, widely known as ResNet architecture \cite{he2016deep}. The residual learning represents an identity mapping through a shortcut connection following simple addition of feature maps of previous layers rendered using  3$\times$3  and 1$\times$1 convolutions. This identity mapping does not incur additional computational overhead and still able to ease degradation problem. In a similar fashion, the MobileNet-V2  uses bottleneck separable convolutions with kernel size 3$\times$3, and inverted residual connection \cite{sandler2018mobilenetv2}. It is a memory-efficient framework suitable for mobile devices.

These backbones are widely used in existing works on diverse image classification problems (e.g., human  activity recognition, object classification, disease prediction, etc.) due to their superior architectural designs \cite{bera2023fine}, \cite{bera2021attend} at reasonable computational cost. Here, these backbones are used for a fair performance comparison with the state-of-the-art methods developed for plant nutrition and  disease classification \cite{singh2020plantdoc}. We have customized the top-layers of base CNNs for adding the GCN module without alerting their inherent layer-wise building blocks, convolutional design such as the kernel-sizes, skip-connections, output feature dimension, and other design parameters. The basic characteristics of these backbone CNNs are briefed in Table  \ref{CNN_Spec}.  The network depth, model size and parameters have been increased  due to the addition of GCN layers upon the base CNN accordingly, evident in Table  \ref{CNN_Spec}.

\begin{table}
\centering
\caption{{Design Specifications of Backbone CNNs and Characteristics of PND-Net} }\label{CNN_Spec}
\begin{tabular}{|c|c|c|c|c|c|c|}
\hline
\multicolumn{5}{|c|}{Backbone CNN Characteristics} & \multicolumn{2}{|c|}{PND-Net properties} \\ \hline
Model Name  & Design Characteristics      & Par (M)  & Depth  & Size (MB) &  Depth &  Size (MB)\\
\hline

Xception \cite{chollet2017xception}  & depth-wise separable convolution & 22.9  & 131   & 85  & 193  &112\\
\hline
ResNet-50 \cite{he2016deep} & residual connections & 25.6 & 174   & 95  &236 & 122\\
\hline

Inception-V3 \cite{szegedy2016rethinking}& Inception module with increased depth & 23.9 & 310   & 92   & 372 &116 \\
\hline
MobileNet-V2 \cite{sandler2018mobilenetv2} & inverted residual and linear bottleneck   & 3.5 & 153   & 10  & 215 & 22\\
\hline

\end{tabular} 
\vspace{- 0.2 cm}
\end{table}

\begin{table}
{
\centering
\caption{{Details of Implementation Specifications} }\label{HW_Spec}
\begin{tabular}{|p{5.5cm}p{4.4cm}p{2.8cm} p{2.2 cm}|}
\hline
  Hardware/Deep Learning Framework  & Training Hyper-parameters & Data augmentation & Time(ms)$/$img \\
\hline
$\square$Tensorflow: 2.13.0, Keras: 2.13.1, Cuda: 12.4, NVIDIA A100 40GB GPU
 $\square$ Intel Core Silver 4316 CPU x86\_64, 2.30 GHz 128 GM RAM  
&
$\square$ Img size: 224$\times$224, Batch: 8 

$\square$ Optimizer: SGD,
 Loss: categorical cross-entropy, 
 Learning rate: 0.007
&
 $\square$ Gaussian noise
 
 $\square$ Random flip, rotation: 20,  scale: 0.20,  translation: 0.20 
 & using ResNet50 
 
 $\square$Train: 15.4    

  $\square$Inference: 5.8   
 \\
\hline
\end{tabular} 
}
\vspace{ -0.3 cm}
\end{table}

Two GCN layers have been used with  ReLU activation, and the feature size is the same as the base CNN's output channel dimension. For example, the size of channel features of ResNet-50, Xception and Inception-V3 is 2048, which is kept the same dimension as GCN's channel feature map. The adjacency matrix is developed considering overall spatial relation among different neighborhood regions as a complete graph. Therefore, each region is related with all other regions even if they are far apart which is helpful in capturing long-distant feature interactions and building a holistic feature representation via a complete graph structure. Batch normalization and a drop-out rate of 0.3 is applied in the overall network design to reduce overfitting. 

\subsubsection{Data Pre-processing and Data Splitting Techniques}
The basic pre-processing technique provided by the Keras applications for each backbone has been applied. It is required  to convert the input images from RGB to BGR, and then each color channel is zero-centered  with respect to the ImageNet dataset, without any scaling. Data augmentation methods such as random  rotation ($\pm$25 degrees), scaling ($\pm$0.25), Gaussian blur, and  random cropping  with 224$\times$224 image-size from the input size of 256$\times$256 are applied on-the-fly for data diversity in image samples.  

We have maintained the same train-test split provided with the datasets e.g., PlantDoc. However, other plant datasets does not provide any specific image distribution. Thus, we have randomly divided the datasets into train and test samples following a 70:30 split ratio which is complied in several works. The details of  image distribution is provided in Table \ref{bl_imgnet}. For cross-validation, we have randomly divided the training samples  into training and validation set with a 4:1 ratio i.e., five-fold cross validation in a disjoint manner, which is a standard techniques adopted in other methods \cite{hameed2022multiclass}. The test set remains unaltered for both evaluation schemes for clear performance comparison. Finally, the average test accuracy of five executions on each dataset has been reported here as the overall performance of the PND-Net.

A summary of the implementation specification indicating the hardware and software environments, training hyper-parameters, data augmentations, and estimated time (milliseconds) of training and inference are specified in Table \ref{HW_Spec}. Our model is trained with a mini-batch size of 12 for 150 epochs and divided by 5 after 100 epochs. However, no other criterion such as early stopping has been followed. The proposed method is developed in  Tensorflow 2.x using Python. 

\subsection{Dataset Description} 

%%%%%%%%%%%%%%
\begin{figure}
\centering
{
    \includegraphics[width=0.32\textwidth, height=3.4 cm]{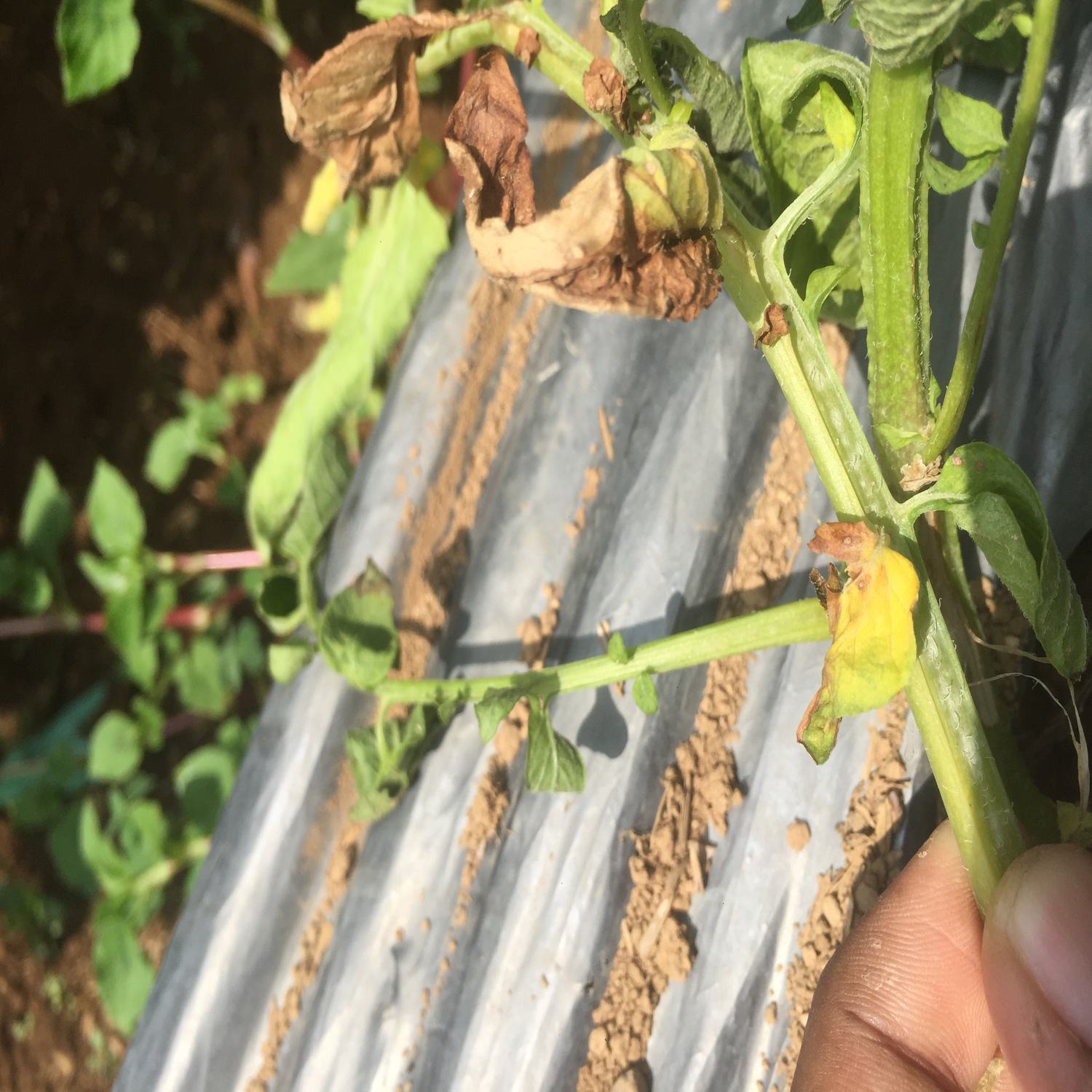}    \hfill
    \includegraphics[width=0.32\textwidth, height=3.4 cm]{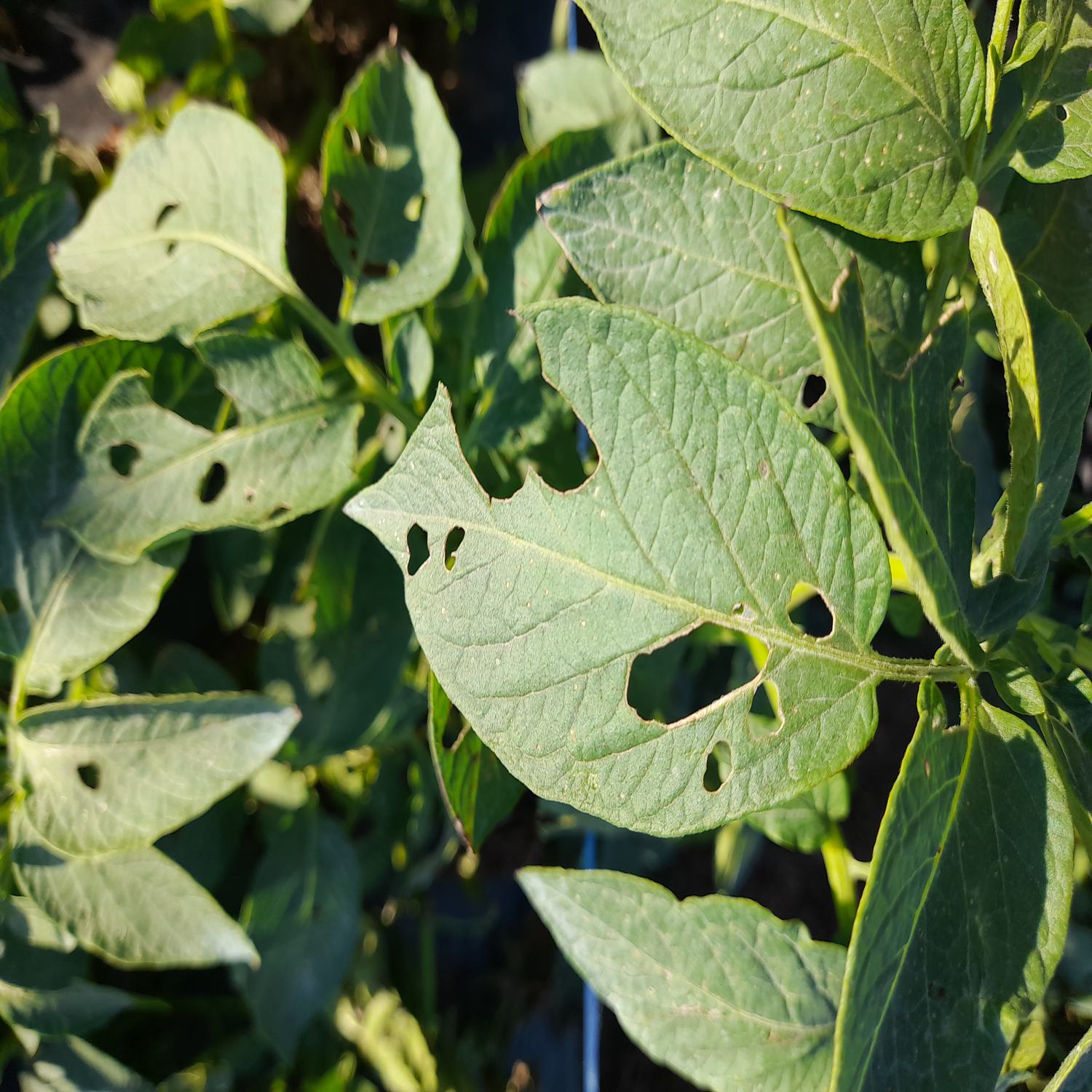}   \hfill  
    \includegraphics[width=0.32\textwidth, height=3.4 cm]{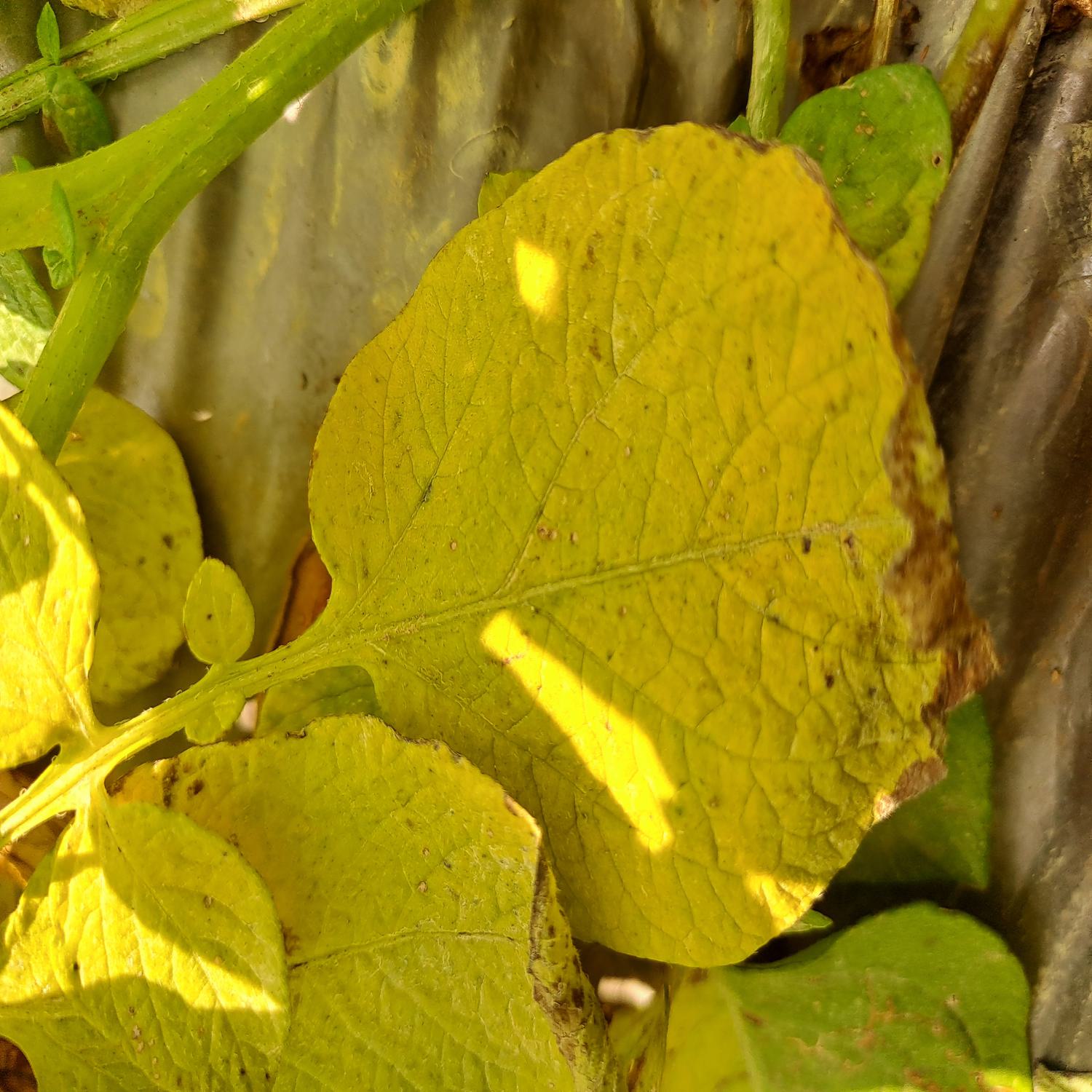}  }
\caption{Sample images of potato diseases infected by bacteria, pest, and Nematodes.}
    \label{Potato}
\end{figure}

\begin{figure}
\centering
{
\includegraphics[width=0.32\textwidth, height=3.4 cm]{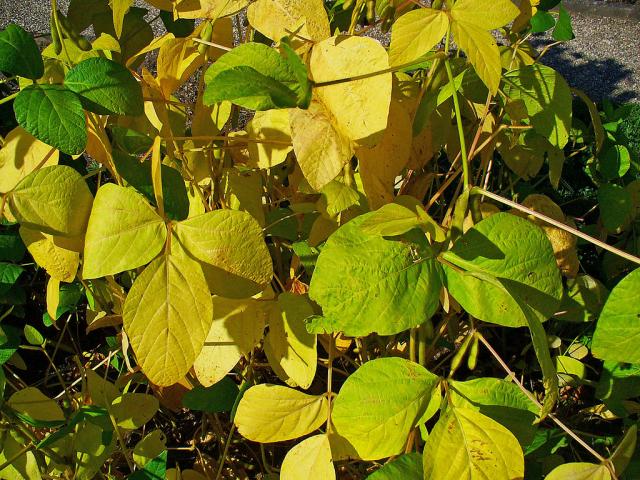}    \hfill
   \includegraphics[width=0.32\textwidth, height=3.4 cm]{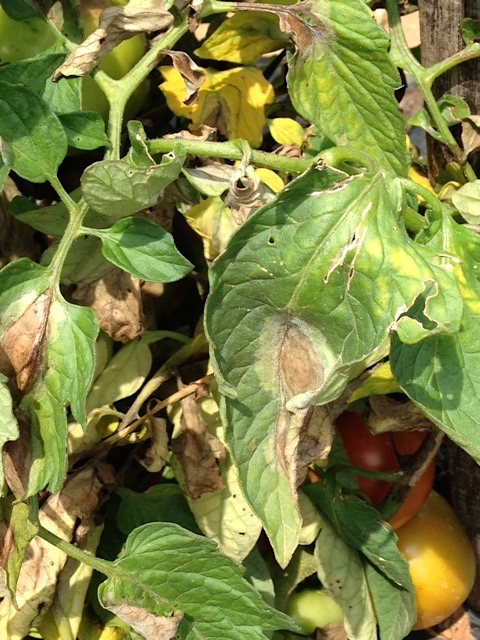} \hfill
      \includegraphics[width=0.32\textwidth, height=3.4 cm]{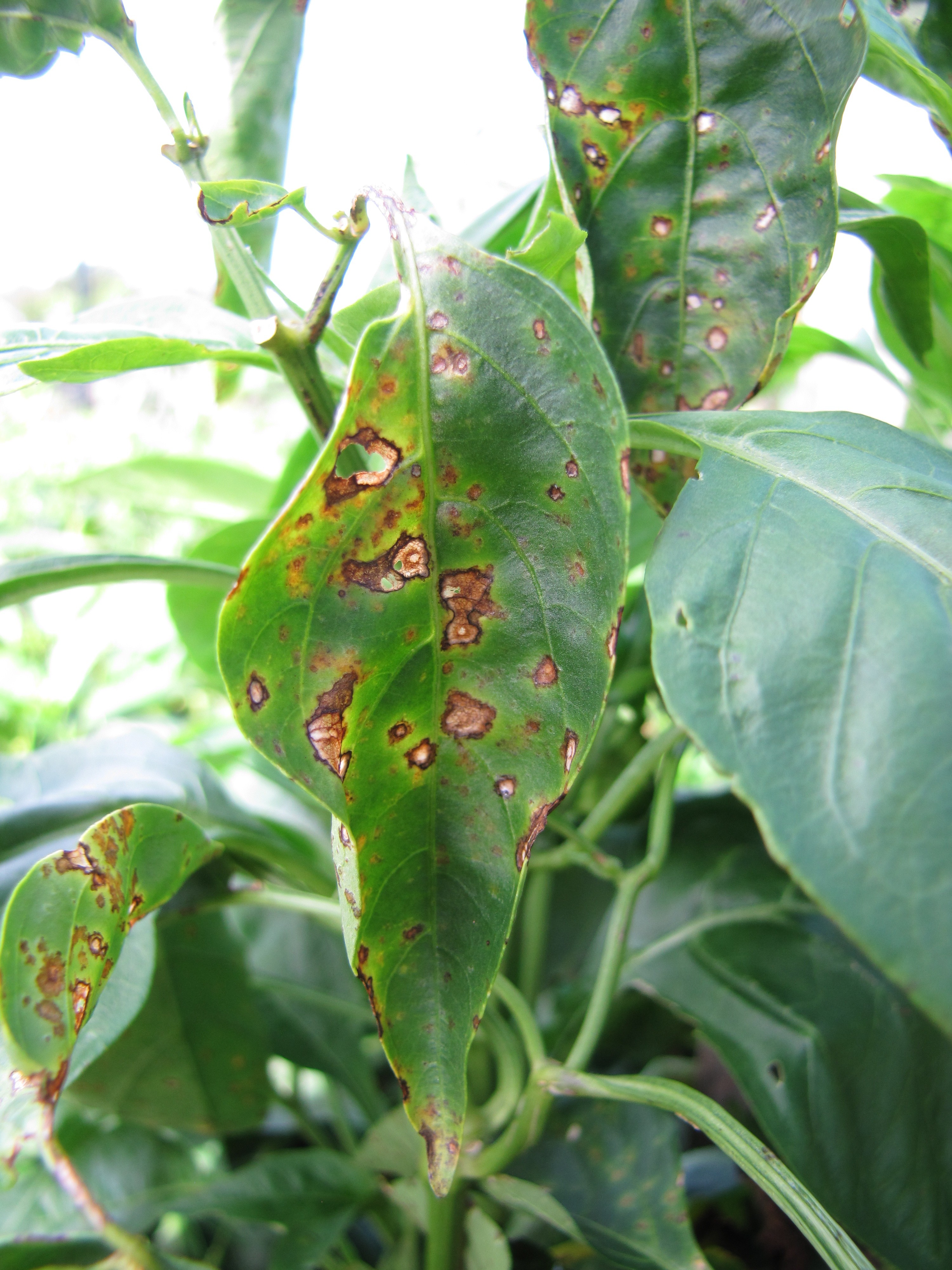} 
   }
\caption{Sample images of infected leaves of soybean, tomato, and bell pepper from the PlantDoc dataset.}
    \label{plantdoc}
\end{figure}

\begin{figure}
\centering
{
    \includegraphics[width=0.32\textwidth, height=3cm]{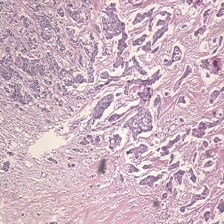}   \hfill
    \includegraphics[width=0.32\textwidth, height=3cm]{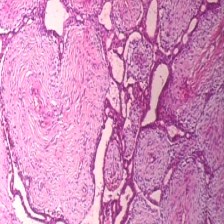}   \hfill 
    \includegraphics[width=0.32\textwidth, height=3cm]{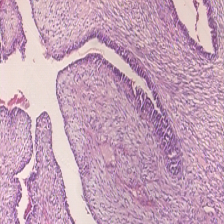}   \hfill 
    }
\caption{Sample images of the BreakHis-40X dataset.}
   \label{Sample_BrkHis}
      \vspace{ -0.2 cm}
\end{figure}

\begin{figure}
\centering
{
    \includegraphics[width=0.32\textwidth, height=3cm]{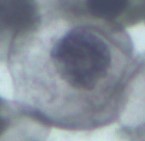}   \hfill
    \includegraphics[width=0.32\textwidth, height=3cm]{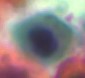}  \hfill
    \includegraphics[width=0.32\textwidth, height=3cm]{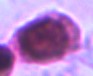}   \hfill 
    }
\caption{Sample images of the SIPaKMeD dataset.}
   \label{Sample_SIPaKMeD}
   \vspace{ -0.4 cm}
\end{figure}

The summary of  four plant datasets used in this work are summarized in Table \ref{bl_imgnet}. These datasets are collected from public repositories such as the Mendeley Data and Kaggle.
\begin{itemize}
 
\item 
The  Banana nutrition deficiency dataset represents healthy samples and the visual symptoms of deficiency of the: Boron, Calcium, Iron,   Magnesium, Potassium,  Sulphur,  and Zinc. The samples of this dataset are shown in Fig. \ref{banana}. More details are provided in Ref \cite{sunitha2023fully}. 

\item 
The  Coffee nutrition deficiency dataset (CoLeaf-DB) \cite{tuesta2023coleaf} represents healthy samples and the deficiency classes are:  Boron, Calcium, Iron, Manganese, Magnesium,   Nitrogen,  Potassium, Phosphorus,    and more deficiencies. The samples of dataset are illustrated in Fig. \ref{coffee}. 

\item 
The  Potato disease classes are: Virus, Phytopthora, Pest, Nematode, Fungi, Bacteria, and healthy. The samples of this  dataset are shown in Fig. \ref{Potato}. The dataset is collected from the Mendeley \cite{shabrina2024novel}  repository. 

\item      
The PlantDoc is a realistic plant disease dataset \cite{singh2020plantdoc}, comprising with different disease classes of Apple, Tomato, Potato, Strawberry,  Soybean, Raspberry, Grapes, Corn, Bell-pepper,  and others.  Examples are shown in Fig. \ref{plantdoc}.

\item The Breast Cancer Histopathology Image Classification (BreakHis) \cite{spanhol2015dataset} dataset with 40X and 100X magnifications contain  8-classes:  adenosis, fibroadenoma, phyllodes tumor, and tubular adenoma; ductal carcinoma, lobular carcinoma, mucinous carcinoma, and papillary carcinoma. The samples of this dataset are exemplified in Fig.  \ref{Sample_BrkHis}.

\item  The SIPaKMeD \cite{plissiti2018sipakmed}, containing 4050 single-cell images, which is useful for classifying  cervical cells in Pap smear images, shown in Fig. \ref{Sample_SIPaKMeD}.  This dataset is categorized into  five  classes based on cytomorphological features. %
        
\end{itemize}

%%%%%%%%%%%%%%

%
\begin{table}
\centering
\caption{{ Dataset summary with the baseline accuracy (\%)} using  different base CNNs only }\label{bl_imgnet}
\begin{tabular}{|c|c|c|c||c|c|c|c|}
\hline
Dataset Name & Class & Train & Test &  Xception  & ResNet-50 &   MobileNet-V2 & Inception-V3  \\
\hline
Banana Nutrition Deficiency  & 8 & 2156 & 920 & 62.50 & 61.53  & 61.74 &62.17  \\
Coffee Nutrition Deficiency & 9 &700 & 300 & 69.25 &  68.24 & 66.55  & 69.93 \\

Potato Disease & 7 & 2010 & 869 & 83.56  &  83.16 & 80.78 & 83.44 \\

PlantDoc Disease & 27 &  2047& 516 & 64.64  & 64.45  &  61.32 & 60.15\\

\hline
BreakHis-40X & 8 &  1400&  600& 86.00  & 87.66  & 84.16  & 85.33 \\
BreakHis-100X & 8 & 1460  & 625 & 81.80  & 83.33  & 80.20 & 82.51\\
SIPaKMeD & 5  & 3550 & 500 & 91.33 & 92.66   & 90.92   & 92.34 \\
\arrayrulecolor{black}\hline
%\hline
\end{tabular}  
\end{table}

\begin{table}
\centering
\caption{{Overall performances (\%) of the proposed PND-Net built upon different standard base CNNs} }\label{AllResult}
\begin{tabular}{|c|c|c|c|c|c|c|}
\hline
Dataset & Base CNN + GCN  & Top-1 Accuracy  & Top-3 Accuracy & Precision &   Recall & F1-score \\
\hline
  & ResNet-50& \textbf{90.00}  &98.34&90.00  &90.00  &90.00 \\
Banana  & Xception & 89.25  &98.27 &90.00  &89.00  &89.00 \\
 & Inception-V3 & 83.77  &98.13 &84.00  &84.00  &84.00 \\
  & MobileNet-V2 & 83.99  &97.80 &84.00  &84.00  &83.00 \\
\hline
 & ResNet-50& 89.52 &  97.00 & 89.00 &89.00    &  89.00\\
Coffee & Xception & \textbf{90.54} &  98.67 & 90.00 &90.00    &  90.00\\

  & Inception-V3 &89.18 & 98.67 & 89.00 &89.00   & 89.00 \\
  & MobileNet-V2 &89.86 &  98.33 & 90.00 &89.00    &  89.00\\
  
\hline
 &ResNet-50 & 94.32 & 99.03 &94.00  & 94.00  & 94.00 \\
Potato &Xception & \textbf{96.18} & 99.42 &96.00  & 96.00   & 96.00 \\
 & Inception-V3 & 96.05 & 99.64 &96.00  & 96.00   & 96.00 \\
 & MobileNet-V2 & 92.59 & 98.68 &93.00  & 93.00   & 93.00 \\
\hline
 & ResNet-50  &84.11 & 98.02  & 85.00 & 84.00  &84.00 \\
 PlantDoc &Xception  & \textbf{84.30} &  98.10 & 85.00 & 84.00& 84.00\\
 &Inception-V3  &81.00 & 98.05  & 81.00 & 81.00 &81.00 \\
 & MobileNet-V2  &80.81 & 97.86  & 81.00 & 81.00  &81.00 \\
\hline

BreakHis 40X & 
 ResNet-50 &\textbf{95.50} & 99.00 &  95.00 & 95.00 & 95.00\\

& Xception  &94.83  & 99.00 & 95.00 &95.00 &95.00 \\
& Inception-V3 & 95.00 &99.00 & 95.00 & 95.00 & 95.00\\
& MobileNet-V2 & 94.00 & 99.00 & 94.00 & 94.00 &94.00 \\
%\hline
\hline
BreakHis 100X 
& ResNet-50 & \textbf{96.79} & 99.00 &  97.00 & 97.00 & 97.00\\

& Xception  & 95.19 & 99.00 & 95.00 &94.00 &94.00 \\

& Inception-V3 &95.67 &99.00 & 96.00 &96.00 & 96.00\\
& MobileNet-V2 &95.83 & 99.00 & 96.00 &96.00 &96.00 \\
\hline

SIPaKMeD  
& ResNet-50 & \textbf{99.18} & 100.00 & 99.00 &99.00 &99.00 \\

& Xception    &98.98 & 100.00 & 99.00 &99.00 &99.00 \\

& Inception-V3 & 98.37&100.00 & 98.00& 98.00&98.00 \\
& MobileNet-V2 &98.17 & 100.00 & 98.00 &98.00 &98.00 \\
\hline
\end{tabular} 
%\vspace{ -0.2 cm}
\end{table}

\subsection{Result Analysis and Performance Comparison}
A summary of the datasets with data distribution, and the baseline accuracy (\%) achieved by aforesaid base CNNs are briefed in Table \ref{bl_imgnet}. The baseline  model is developed using the pre-trained CNN backbones with ImageNet weights. A backbone CNN extracts  the base output feature map which  is   pooled  by a global average pooling layer and classified with a softmax layer.  Four backbone CNNs with different design characteristics  have used for generalizing our proposed method. The baseline accuracies are reasonable and consistent across various datasets, evident in Table \ref{bl_imgnet}.

Two different evaluation strategies i.e., general classification and $k$-fold cross validation ($k=5$) have been experimented. An average performance has been estimated from multiple executions on each dataset and reported here. 
The top-1 accuracies (\%) of the proposed PND-Net comprising  two-GCN layers with the feature dimension 2048, included on the top of different backbone CNNs, are given in Table \ref{AllResult}.  The overall performance of the PND-Net on all  datasets significantly improved over the baselines. Clearly, it shows the efficiency of the proposed method. In addition, the PND-Net model has been tested with five-fold cross validation for a robust performance analysis (Sec \ref{5_foldCV}). These cross-validation results (Table \ref{5fold_potato}-\ref{kfold2}) on each dataset could be considered as the benchmark performances using several metrics. Our method has driven the state-of-the-art performances on these datasets for plant disease and nutrition deficiency recognition. 

\begin{table}
{\centering
\caption{{The performance of PND-Net on the potato disease dataset using five-fold cross validation}} \label{5fold_potato}
\begin{tabular}{|c|c|c|c|c|c|c|c|c|c|c|}
\hline
\multirow{2}{*}{k-Fold} & \multicolumn{5}{|c|}{PND-Net using ResNet-50 base } &  \multicolumn{5}{|c|}{ PND-Net using Xception base}   \\
\cline{2-11}
 & Val Acc & Test Acc & Prec. & Recall  & F1-score & Val Acc & Test Acc & Prec. & Recall  & F1-score \\ \hline

Fold-1 & 96.46 & 95.48  & 96.00  &  95.00 & 96.00  &  92.67  & 91.32 & 91.00  &91.00 & 91.00 \\
 
Fold-2 & 95.70 & 95.27  & 95.00 & 95.00  & 95.00 & 91.66 & 91.31& 91.00  & 91.00  &  91.00 \\

Fold-3 & 95.20 & 94.66  & 95.00 & 95.00  & 95.00 & 94.39  & 92.13 & 92.00  & 92.00 & 92.00\\

%\hline
Fold-4 & 94.95 & 93.27  & 93.00 & 93.00  & 93.00 &  93.90 & 91.55& 91.00  &91.00 & 91.00\\

Fold-5 & 95.70 &  95.20 & 95.00& 95.00  & 95.00 & 94.91  & 92.82& 93.00  &93.00 & 93.00 \\
\hline

Avg &  95.60	& 94.78	&94.80	&94.60	&94.80	&93.51	&91.83	&91.60	&91.60	&91.60 \\

\hline
\end{tabular} 
}
\end{table}

An experimental study has been carried out on two more public datasets for human medical image analysis. The BreakHis with 40X and 100X magnifications \cite{spanhol2015dataset}  and SIPaKMeD \cite{plissiti2018sipakmed} datasets have been evaluated for generalization.   
The  SIPaKMeD dataset \cite{plissiti2018sipakmed} is useful for classifying  cervical cells in pap smear images, illustrated in Fig. \ref{Sample_SIPaKMeD}.  This dataset is categorized  into  five  classes based on cytomorphological features using the proposed PND-Net. The conventional classification results  are given in Table \ref{AllResult}, and the performances of  cross validations are provided in Table \ref{A5fold_BHis40} and \ref{kfold2}.

\begin{table}

\centering
\caption{{The performance of PND-Net on the BreakHis-40X dataset using five-fold cross validation} }\label{A5fold_BHis40}
\begin{tabular}{|c|c|c|c|c|c|c|c|c|c|c|}
\hline
\multirow{2}{*}{k-Fold} & \multicolumn{5}{|c|}{PND-Net using ResNet-50 backbone} &  \multicolumn{5}{|c|}{PND-Net using Xception backbone}   \\
%\hline
\cline{2-11}
 & Val Acc & Test Acc & Prec. & Recall  & F1-score & Val Acc & Test Acc & Prec. & Recall  & F1-score \\ \hline

Fold-1 & 98.45 & 97.25  & 97.00 & 97.00  & 97.00 & 96.79  & 96.10 & 96.00  & 96.00 & 96.00\\
 
%\hline
Fold-2 & 97.81 & 96.70  & 96.00 & 96.00  & 96.00 & 97.81  & 97.20 & 97.00 & 97.00 & 97.00\\

%\hline
Fold-3 & 98.68 & 97.30  & 97.00 & 97.00  & 97.00 & 98.12  & 97.75 & 97.00  &97.00 & 97.00\\

%\hline
Fold-4 & 98.75 & 97.50  & 97.00 & 98.00  & 98.00& 97.18  & 96.30& 97.00  &96.00 & 96.00\\
%\hline

Fold-5 &98.07 & 96.75  & 97.00 & 96.00  & 97.00 & 97.50  & 96.15 & 96.00  &96.00 & 96.00\\
\hline

Avg & 98.35 & 97.10  & 96.80 & 96.80  & 97.00 & 97.48  & 96.70 & 96.60  &96.40 & 96.40\\
\hline
\end{tabular} 
%\vspace{-0.5 cm}
\end{table}

\subsubsection{Five-fold Cross Validation Experiments} \label{5_foldCV}
The 5-fold cross-validation experiments on  various datasets have been conducted for  evaluating the performance of PND-Net using the ResNet-50 and Xception backbones, and the results are given in  Table \ref{5fold_potato}, \ref{A5fold_BHis40}, and \ref{kfold2}. The  actual train set is divided into five disjoint subsets of images for each dataset. In each experiment, four out of five subsets are used for training and the remaining one is validated independently. Finally, the average validation result of five folds is reported.

The  results of five-fold cross validation on potato leaf disease dataset are provided in  Table \ref{5fold_potato}. The numbers of potato leaf images in each fold for training, validation, and testing are  1608, 402, and 869, respectively. The results using different metrics are computed  and the last row implies an average performance of cross validation on this dataset.

Likewise, the  performance five-fold validation on the BreakHis-40X dataset has been presented in  Table \ref{A5fold_BHis40}. In this experiment, the number of training samples in each fold is 1280 images, and validation set containing remaining 320 images. The test set contains 400 images which  remains the same as used in aforesaid other experiments. Each of the five-fold experiment has been  validated and tested on the test set. Lastly, an average result of five-fold cross validation has been computed, and given in the last row of  Table \ref{A5fold_BHis40}. 

A similar experimental set-up of five-fold cross validation has been followed for other datasets. The average performances of PND-Net on these datasets are provided in Table \ref{kfold2}. The average cross-validation results are better than the conventional classification approach on the potato disease (ResNet-50: 94.48\%) and BreakHis-40X (ResNet-50: 97.10\%) datasets. The reason could be the variations in the  validation set in each fold  enhances the learning capacity of model due to training data diversity. As a result, improved performances have been achieved on diverse datasets. The results are  consistent with the results of conventional classification method on other datasets as described above. The overall performances  on different datasets validates the generalization capability of the proposed PND-Net.  

\begin{table}
\centering
\caption{{Five-fold cross validation and test accuracy (\%) of PND-Net on different datasets} } \label{kfold2}
\begin{tabular}{|c|c|c|c|c|c|c|c|c|c|c|}
\hline
\multirow{2}{*}{CNN} &   \multicolumn{2}{|c|}{Banana}  & \multicolumn{2}{|c|}{Coffee} &  \multicolumn{2}{|c|}{PlantDoc} & \multicolumn{2}{|c|}{ BreakHis-100X}  & \multicolumn{2}{|c|}{ Cell PAP} \\
%\hline
\cline{2-9}
 & Val Ac & Test Ac & Val Ac  & Test Ac     & Val Ac & Test Ac & Val Ac  & Test Ac  & Val Ac  & Test Ac  \\

\hline
Xception & 91.36 & 88.25  & 91.90 & 87.84 & 86.71 & 84.57 &95.97 & 94.63 &99.10 & 97.66\\

ResNet-50 & 93.72 & 89.40  & 94.30 & 90.88 &  85.93 & 83.78 & 97.21& 96.11 &99.70 & 98.92\\
\hline
\end{tabular} 
\end{table}

\begin{table}
\centering
\caption{{Model Parameters (Millions) of PND-Net including base CNNs and GCNs with the feature size = 1024 and 2048, specified within parenthesis} }\label{model_param}
\begin{tabular}{|c|c|c|c|c|}
\hline
Backbone CNN/PND-Net Method &   Xception  & ResNet-50 &   MobileNet-V2 &  Inception-V3  \\
\hline

{CNN Baseline} & 22.88 & 25.61 & 3.50 & 23.87\\
{PND-Net (GCN feature=1024)} &26.13 & 28.86  &6.75   &27.08  \\
{PND-Net (GCN feature=2048)} &37.66 & 40.41   & 17.51 & 38.40 \\
\hline
\end{tabular} 
\end{table}

 \begin{figure}
\centering
{
    \includegraphics[width=0.85\textwidth, height=7.0 cm]{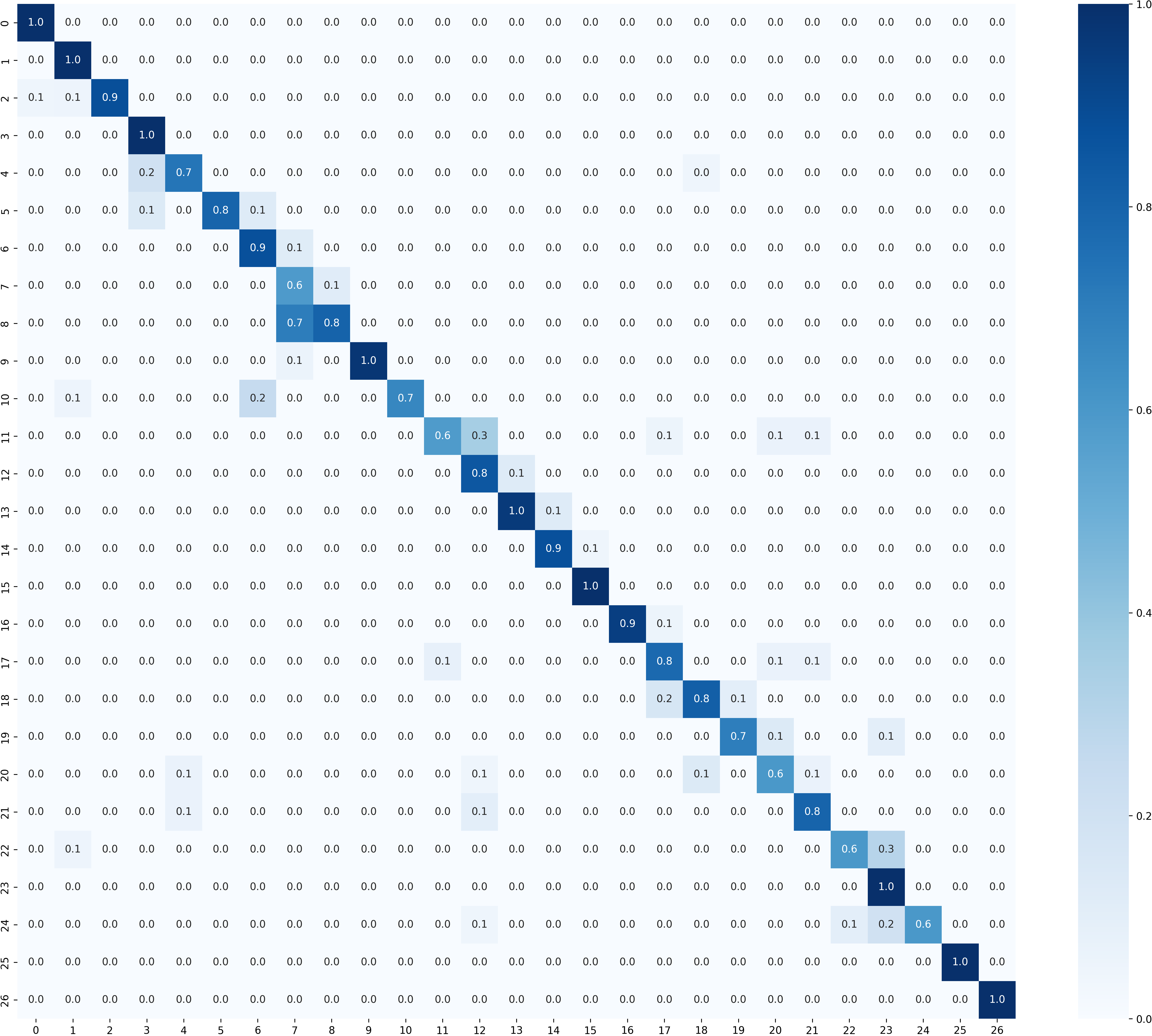}  
        
    \vspace{0.3 cm}
                
    \includegraphics[width=0.32\textwidth, height=3.2 cm]
        {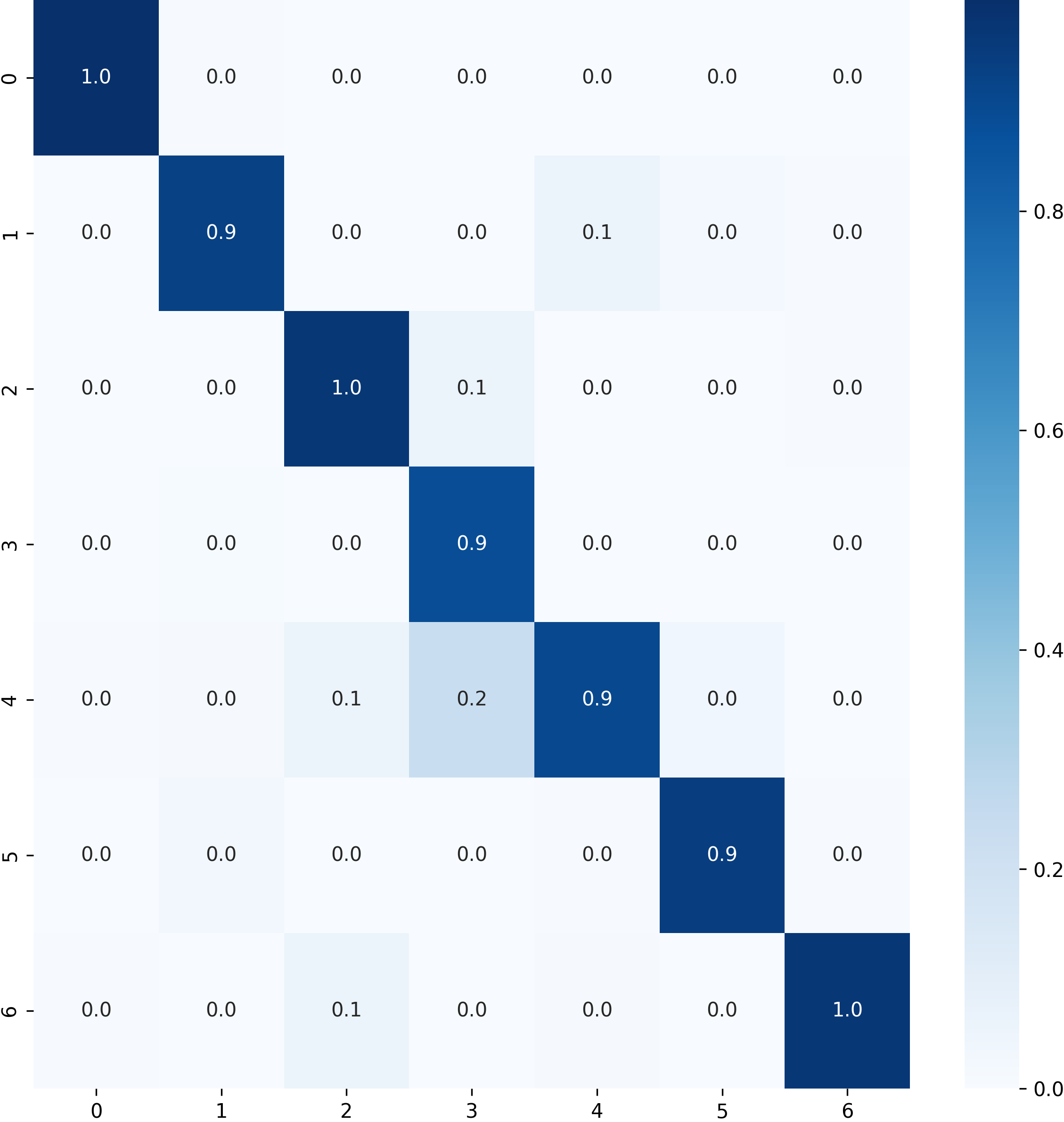}   \hfill 
    \includegraphics[width=0.32\textwidth, height=3.2cm]{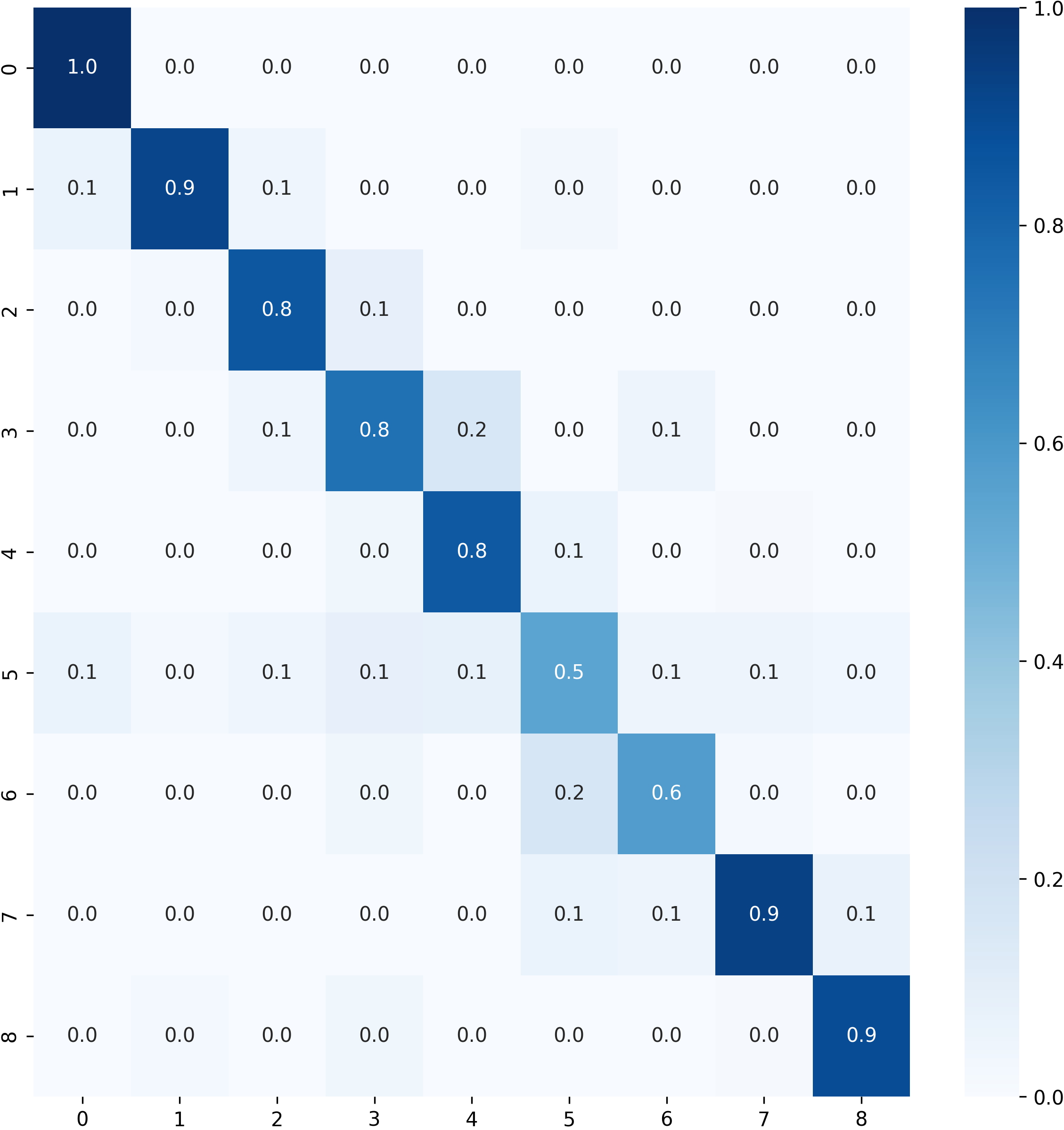}   \hfill 
    \includegraphics[width=0.32\textwidth, height=3.2cm]{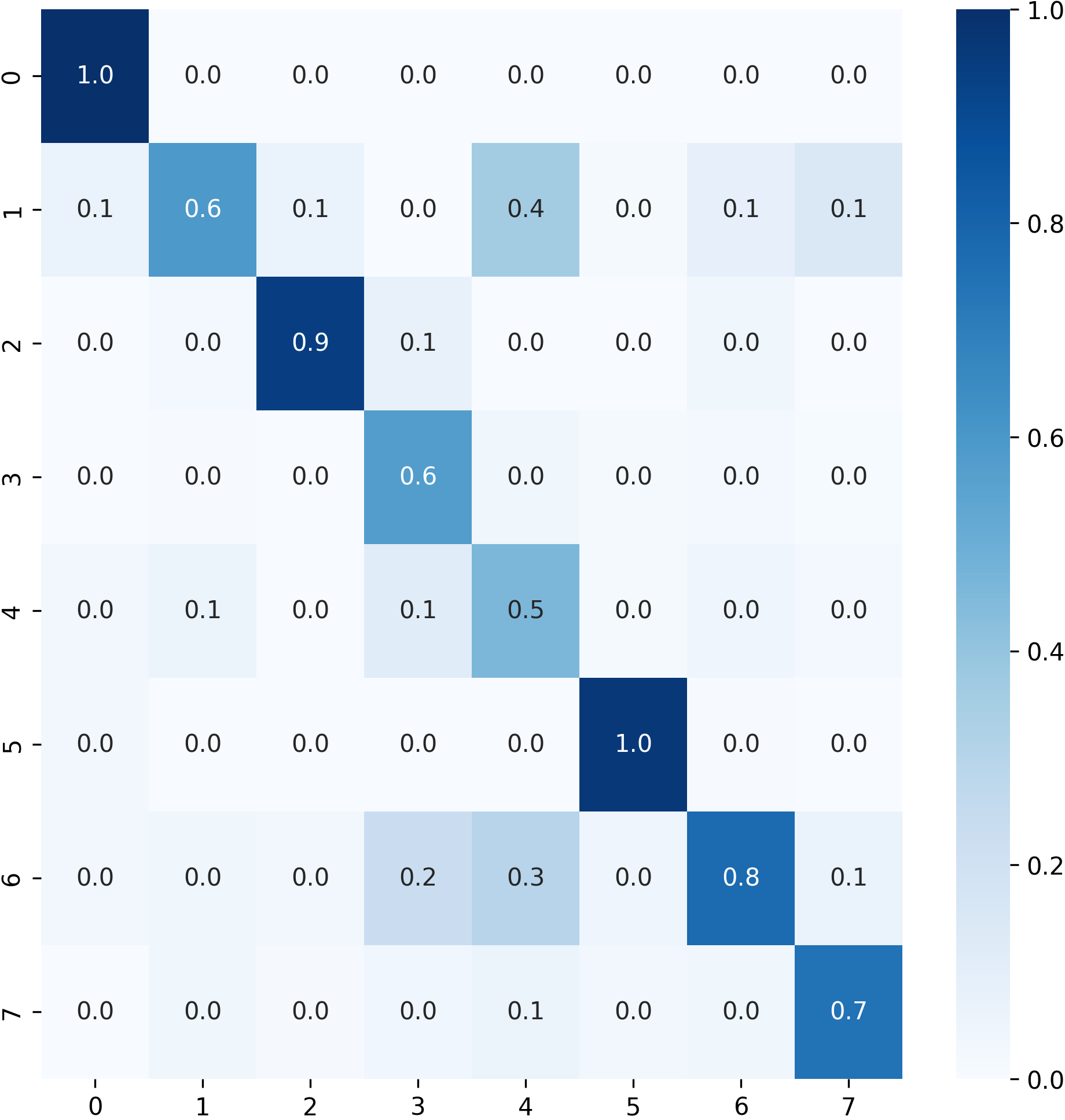}   \hfill 
           
    }
\caption{Confusion matrices  have been computed using  the proposed PND-Net with ResNet-50 backbone on: (a) top-row: PlantDoc; (b) bottom-row: Potato, Coffee, and Banana datasets.}
   \label{cm}
\vspace{-0.3 cm}  
\end{figure}

\begin{figure}
\centering
{
        \includegraphics[width=0.48\textwidth, height=3.5 cm]{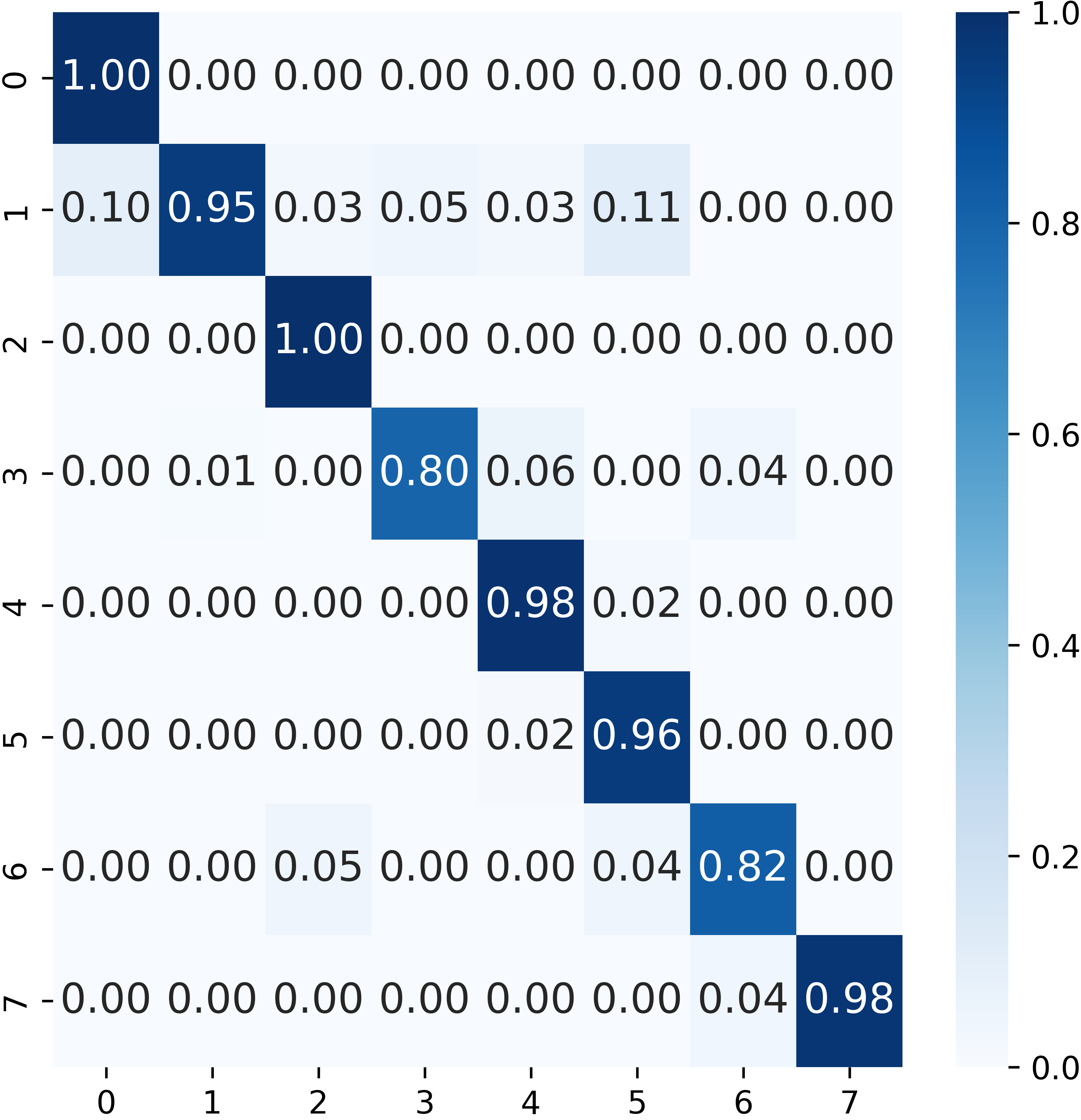}  \hfill \hspace{ 1 cm}
           \includegraphics[width=0.40\textwidth, height=3.5cm]{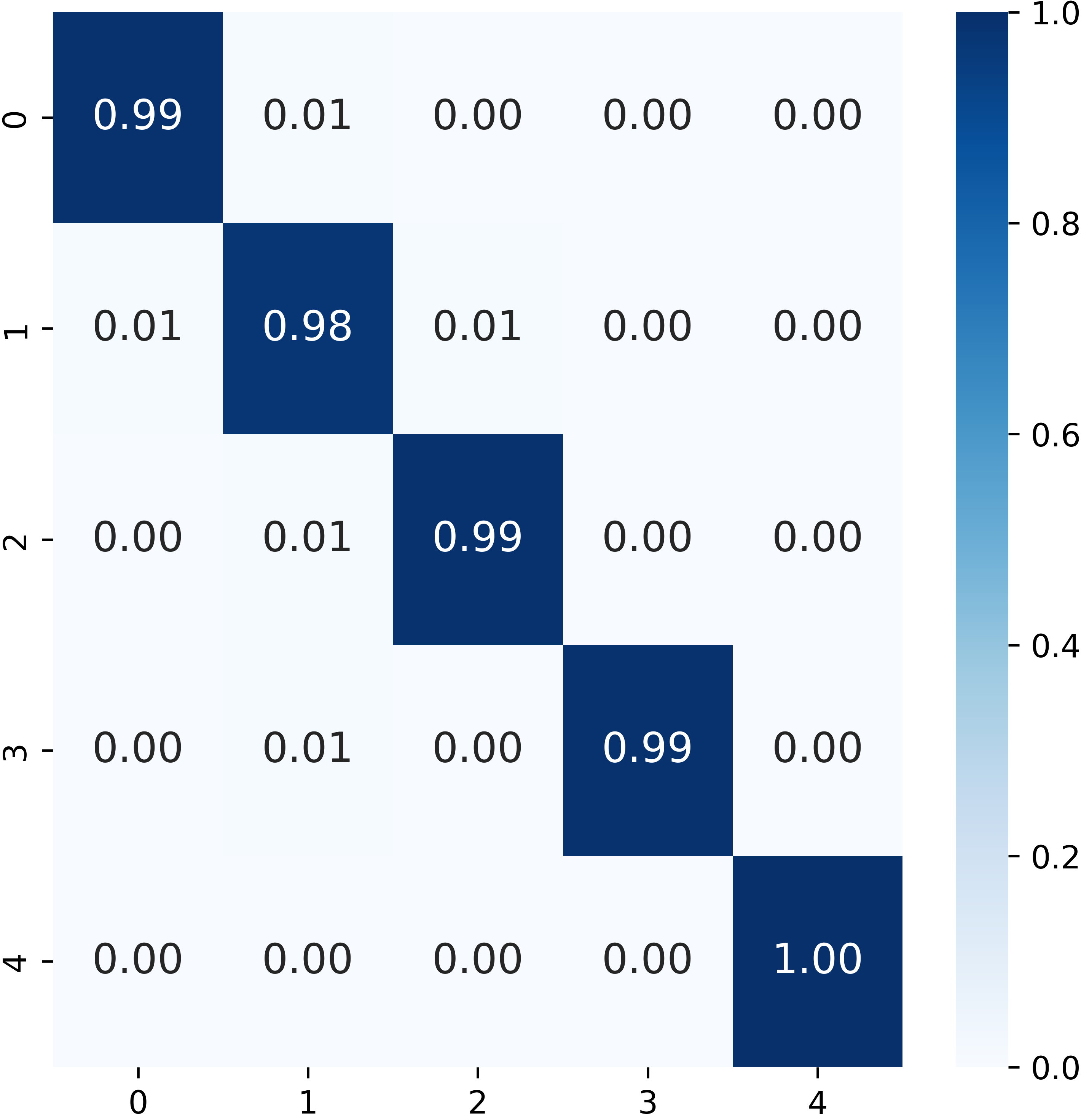}   \hfill 
    }
\caption{Confusion Matrix on the BreakHis-40X dataset  (left) and Smear PAP Cell  dataset (right) using the proposed PND-Net built upon the Xception backbone.
}
   \label{CM_Brk}
   \vspace{-0.3cm}
\end{figure}

\begin{figure}
\centering
{
        \includegraphics[width=0.48\textwidth, height=4.5 cm]{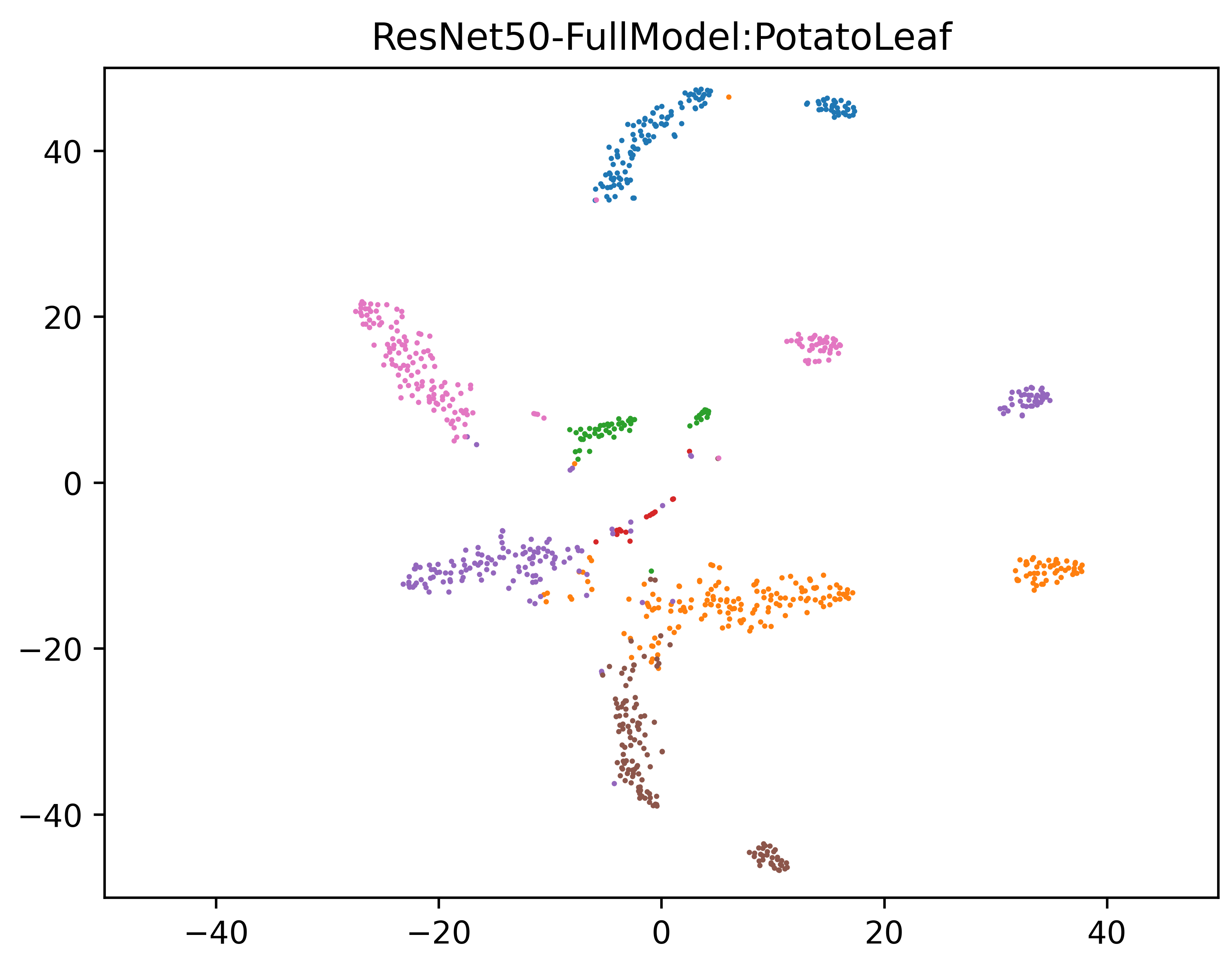}  
           \includegraphics[width=0.48\textwidth, height=4.5 cm ]{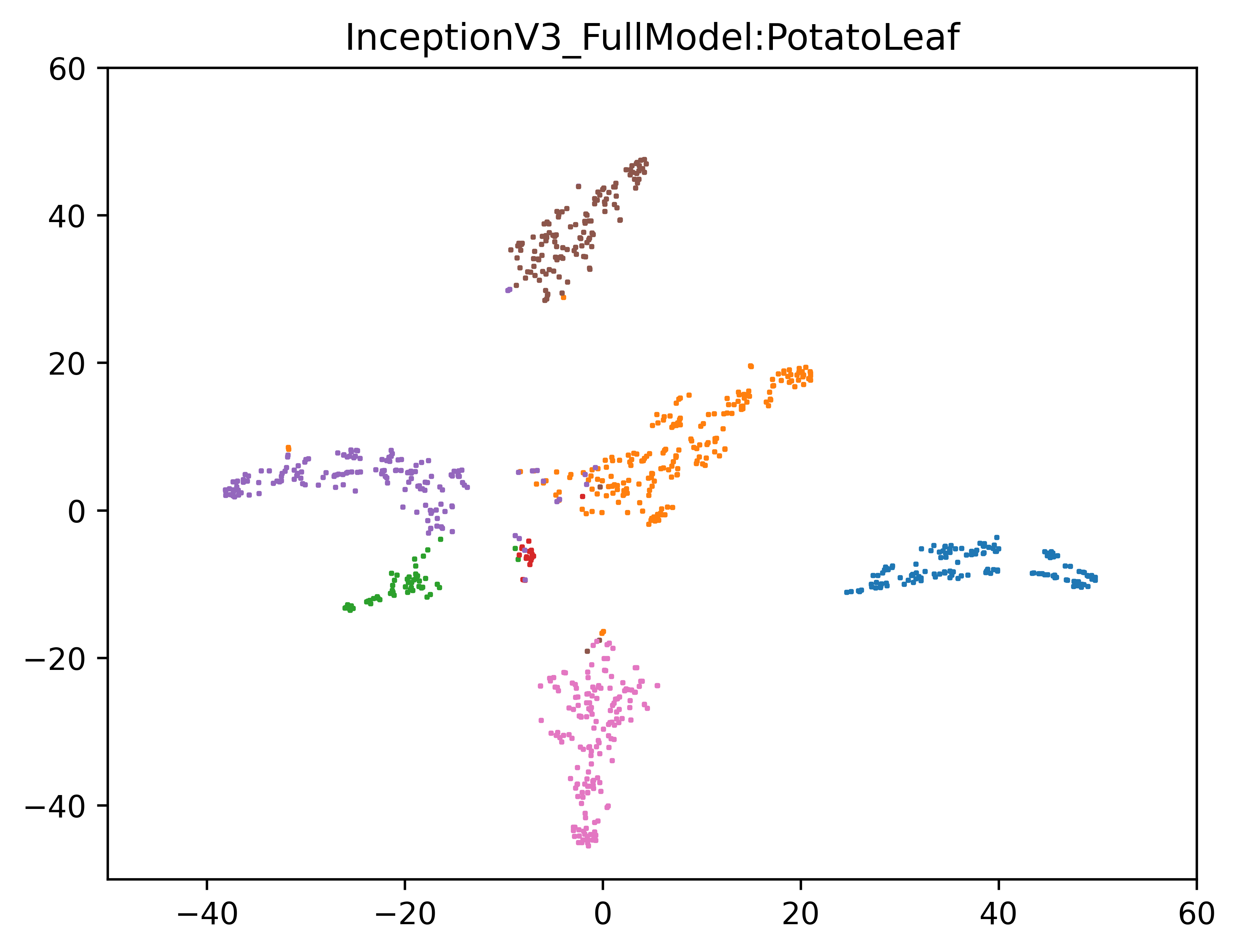}   \hfill 
    }
\caption{ The t-SNE plots on the Potato leaf dataset using  PND-Net with ResNet-50 (left) and Inception-V3 (right).}
 %\vspace{- 1 cm}
   \label{tsne}
\end{figure}

\begin{figure}
\centering
{
    \includegraphics[width=0.24\textwidth, height=3.2 cm]{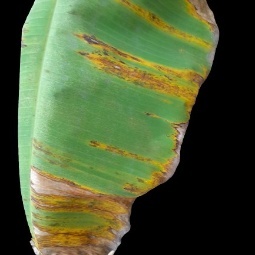}  \hfill 
    \includegraphics[width=0.24\textwidth, height=3.2 cm]{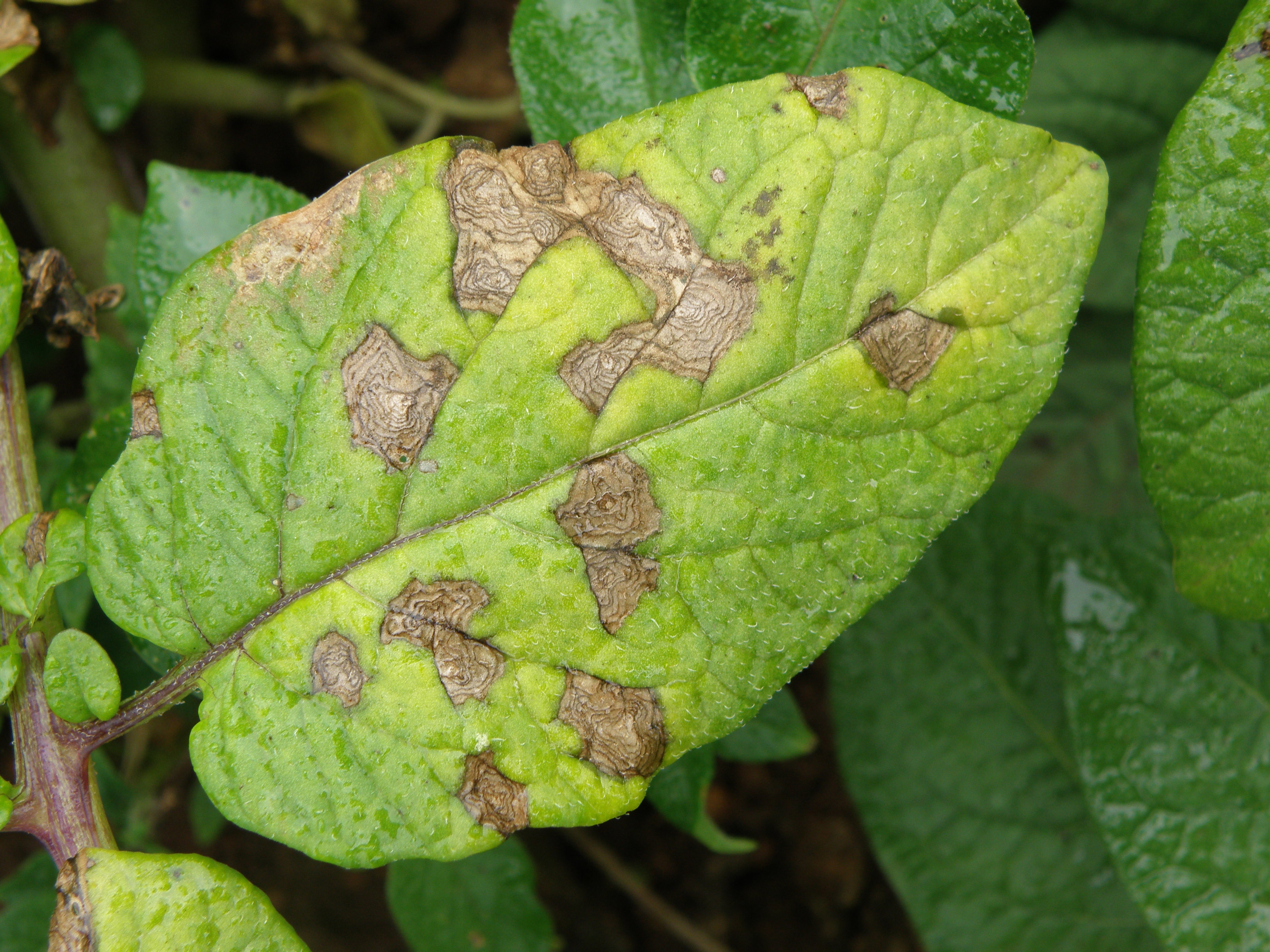}\hfill 
    \includegraphics[width=0.24\textwidth, height=3.2 cm]{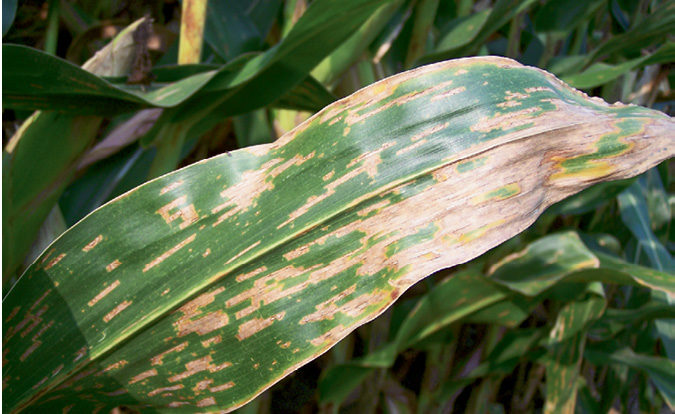}\hfill 
    \includegraphics[width=0.24\textwidth, height=3.2 cm]{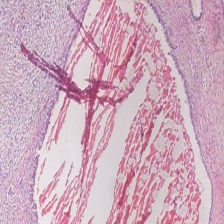}  \\
\vspace{0.2 cm}
        
    \includegraphics[width=0.24\textwidth, height=3.2 cm ]{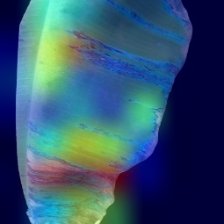}    \hspace{0.07cm}
    \includegraphics[width=0.24\textwidth, height=3.2 cm ]{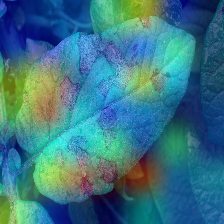}   \hfill 
    \includegraphics[width=0.24\textwidth, height=3.2 cm ]{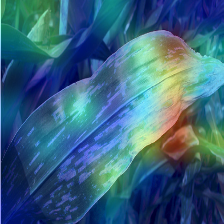}   \hfill 
    \includegraphics[width=0.24\textwidth, height=3.2 cm ]{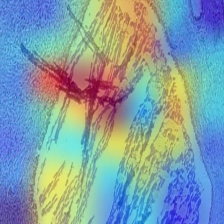}   
    }
\caption{The Grad-CAM output of various datasets are shown,  from left to right: nutrition deficiency,  potato and corn diseases, and breast cancer. The top-row shows an original image and its corresponding Grad-CAM image is shown in the bottom row.}
 %\vspace{- 1 cm}
   \label{grad_cam}
\end{figure}

\subsubsection{Model Complexity and Visualization}
The model parameters are computed in millions, as provided in  Table \ref{model_param}. The model parameters have been estimated for three cases: (a) baseline i.e., the backbone CNN only; and  the output feature dimension of GCN layers is (b) 1024 and (c) 2028.  
{ An average computational time of PND-Net using ResNet-50 has been estimated. The training time is 15.4 ms per image, and inference time is 5.8 ms per image,  and model size is 122MB (given in Table \ref{HW_Spec})}. 
The confusion matrices on these four plant datasets are shown in Fig. \ref{cm}, indicating  an overall performance using ResNet-50. Also, the feature map distributions are clearly shown in different clusters in the t-SNE diagrams \cite{van2014accelerating}  represented with two backbone models on the potato leaf dataset, shown in Fig. \ref{tsne}.  The gradient-weighted class activation mapping (Grad-CAM) \cite{selvaraju2017grad}  has been illustrated in Fig. \ref{grad_cam}  for visual explanations  which clearly show the discriminative regions of different images.

\subsubsection{Performance Comparison}
The highest accuracy on Banana nutrition classification was 78.76\% and 87.89\% using the raw dataset and an augmented version of the original dataset \cite{han2023banana}. In contrast, our method has attained 84.0\% using lightweight MobileNet-V2 and the best 90.0\% using ResNet-50 on the raw dataset, implying a significant improvement in accuracy on this dataset. 

The performances of PND-Net on the Coleaf-DB (Coffee dataset) are very similar, and the best accuracy (90.54\%) is attained by the Xception. The  differences of performances with other base CNNs are very small, implying a consistent performance. The elementary result using ResNet-50 reported on this  recent public dataset is 87.75\% \cite{tuesta2023coleaf}. Thus, our method has set new benchmark results on Coleaf-DB for further enhancement in the future.
Likewise, the Potato Leaf Disease dataset is a new one \cite{shabrina2024novel}, collected from Mendeley data source. We are the first to provide in-depth results on this realistic dataset  acquired in an uncontrolled environment.

A  deep learning method has attained 81.53\% accuracy using Xception and 78.34\% accuracy using Inception-V3 backbone on the PlantDoc dataset \cite{ahmad2023toward}. In contrast, our PND-Net has attained 84.30\% accuracy using Xception and 81.0\% using Inception-V3, respectively. It evinces that PND-Net is more effective in discriminating plant diseases compared to the best reported existing methods. Clearly, the proposed graph-based network (PND-Net) is capable of distinguishing  different types of nutrition deficiencies and plant diseases with a higher success rate in real-world  public datasets.

The BreakHis dataset has  been studied for categorizing into 4-classes and binary classification in several existing works. However, we have compared it with the works of classifying into 8 categories at the image-level for a fair comparison.  The top-1 accuracy  attained using Xception  is 94.83\%, whereas  the state-of-the-art accuracy on this dataset is 93.40$\pm$1.8\% achieved using a hybrid harmonization technique \cite{abdallah2023enhancing}. The accuracy reported is 92.8 $\pm$2.1\% using a class structure-based deep CNN \cite{han2017breast}. The cross-validation results (ResNet-50: 97.10\%) are improved over  existing methods.

Several deep learning  methods have been experimented with the SIPaKMeD dataset.
A CNN-based method achieved 95.35 $\pm$ 0.42\% accuracy \cite{plissiti2018sipakmed}, a PCA-based technique obtained 97.87\%  accuracy for 5-class classification \cite{basak2021cervical},  98.30\% using Xception \cite{mohammed2021single}, and 98.26\%  using  DarkNet-based exemplar pyramid deep  model \cite{yaman2022exemplar}. A GCN-based method has reported  98.37$\pm$ 0.57\% accuracy \cite{shi2021cervical}. A few more comparative results  have been studied in Ref  \cite{jiang2023deep}. In contrast, our method has achieved 98.98 $\pm$ 0.20\% accuracy and  99.10\% test accuracy with cross validation using Xception backbone on this dataset. The confusion matrices on both human disease datasets are shown in Fig. \ref{CM_Brk}.
Overall rigorous experimental results imply that the proposed method has achieved state-of-the-art performances on different types of datasets representing plant nutrition deficiency, plant disease, and human disease classification.

\begin{table}
\centering
\caption{{Ablation Study: Top-1 accuracy (\%) of baseline CNNs in addition to region pooling} }\label{Abln2_RoI}
\begin{tabular}{|c|c|c|c|c|}
\hline
Dataset &   Xception  & ResNet-50 &   MobileNet-V2 & Inception-V3  \\
\hline
Banana  &72.73 & 73.81  & 66.70  & 72.30\\
\hline
Coffee  & 81.73 & 70.95  & 78.37 & 79.39 \\
\hline
Potato & 85.87  & 84.72 &84.60 &  84.38\\
\hline
PlantDoc & 75.85 & 74.21  & 75.97  & 77.34 \\ 
\hline

\end{tabular} 
\vspace{ -0.2 cm}
\end{table}

\subsection{Ablation Study}

An in-depth ablation study has been carried out to observe the efficacy of key components of the  PND-Net. Firstly, the significance of computing different local regions is studied. These fixed-size regional descriptors are combined to create for a holistic representation of feature maps over the baseline features. Notably, the region pooling technique has  improved overall performances on all the datasets, e.g., the gain is more than 12\% on the Banana nutrition deficiency dataset using ResNet-50 backbone. The results of this study are provided in Table \ref{Abln2_RoI}.

Afterward, a component-level study has been evaluated by removing a module from the proposed PND-Net to observe the influence of the key component in  performance.  
An ablation study depicting the significance of spatial pyramid pooling (SPP) layer has been conducted, and the results are shown in Table \ref{Abln2_SPP}.
As the selection of discriminatory information at multiple pyramidal structures has been avoided, the model might overlook finer details which could have been captured at multiple scales by the SPP layer. It causes an obvious degradation of the capacity of network architecture, which is evident from the performances. Thus, capturing multi-scale features is useful to select relevant features for effective learning of plant health conditions.

\begin{table}
\centering
\caption{Ablation Study: Top-1 Accuracy (\%) except SPP layer in the PND-Net architecture  }\label{Abln2_SPP}
\begin{tabular}{|c|c|c|c|c|}
\hline
Dataset &   Xception  & ResNet-50 &   MobileNet-V2 & Inception-V3  \\
\hline
Banana  &79.31 & 80.92  & 75.32 & 77.90\\
\hline
Coffee  &  86.48& 86.14 & 83.10 & 87.16\\
\hline
Potato &  95.94 & 88.65 &92.12  & 93.40 \\
\hline
PlantDoc & 81.83 & 76.56  & 79.68 & 80.66 \\
\hline

\end{tabular} 
\end{table}
\begin{table}
\centering
\caption{{Ablation Study: Top-1 accuracy except the GCN layers} }\label{Abln1}
\begin{tabular}{|c|c|c|c|c|}
\hline
Dataset &   Xception  & ResNet-50 &   MobileNet-V2 & Inception-V3  \\
\hline
Banana  &81.46  &  78.23& 82.11 & 78.66\\
\hline
Coffee  &  88.85& 87.83 & 86.48 & 87.50\\
\hline
Potato & 93.17  & 92.59  & 92.82  & 92.12 \\
\hline
PlantDoc &  81.05 &  78.90 & 80.46 & 80.66 \\
\hline
\end{tabular} 
\vspace{-0.5 cm}
\end{table}

\begin{table}
\centering
\caption{{Ablation Study: Top-1 Accuracy using One GCN layer (\%)} }\label{Abln1L}
\begin{tabular}{|c|c|c|c|c|}
\hline
Dataset &   Xception  & ResNet-50 &   MobileNet-V2 & Inception-V3  \\
\hline
Banana  & 86.00 &78.33  & 70.79 & 81.03\\
\hline
Coffee &89.19 & 89.86 & 89.78   & 87.83 \\
\hline
Potato &94.79 & 94.10  &   93.98 & 92.84 \\
\hline
PlantDoc & 80.27  & 79.68  &  77.34 &80.07 \\
\hline
\end{tabular} 
\end{table}

\begin{table}
\centering
\caption{{Ablation Study: Top-1 Accuracy (\%) with feature dimension 1024 in GCN layers of PND-Net} }\label{Abln_GCN1024}
\begin{tabular}{|c|c|c|c|c|}
\hline
Dataset &   Xception  & ResNet-50 &   MobileNet-V2 & Inception-V3  \\
\hline
Banana  &87.04 & 85.74 &76.42 & 82.45\\
\hline
Coffee  &85.66 &84.00 & 83.33& 86.66 \\
\hline
Potato &  95.02 & 93.40 & 95.13 & 91.20\\
\hline
PlantDoc &83.91  & 79.65& 75.97 & 77.90 \\
\hline
\end{tabular} 
\end{table}

Next, the efficacious GCN modules are excluded from the network architecture, and then, experiments have been conducted with  regional features selected by our composite pooling modules (i.e., regions + SPP) from upsampled high-level deep features of a base CNN. The results are provided in Table \ref{Abln1}. 

\begin{figure}
\centering
{
\subfigure[] 
            {
    \includegraphics[width=0.48\textwidth, height=4 cm]{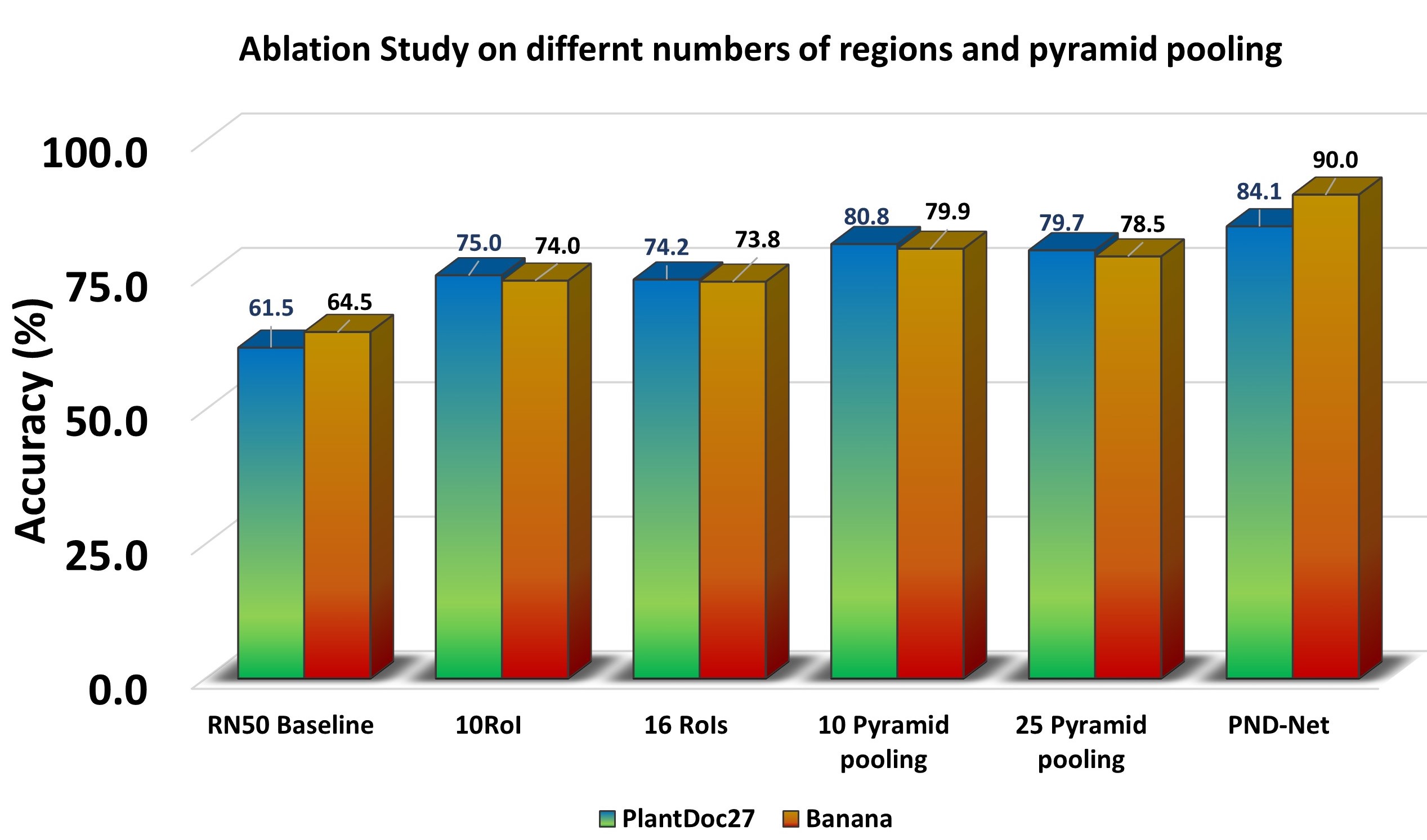} } \hfill
   \subfigure[]{ \includegraphics[width=0.48\textwidth, height=4 cm]{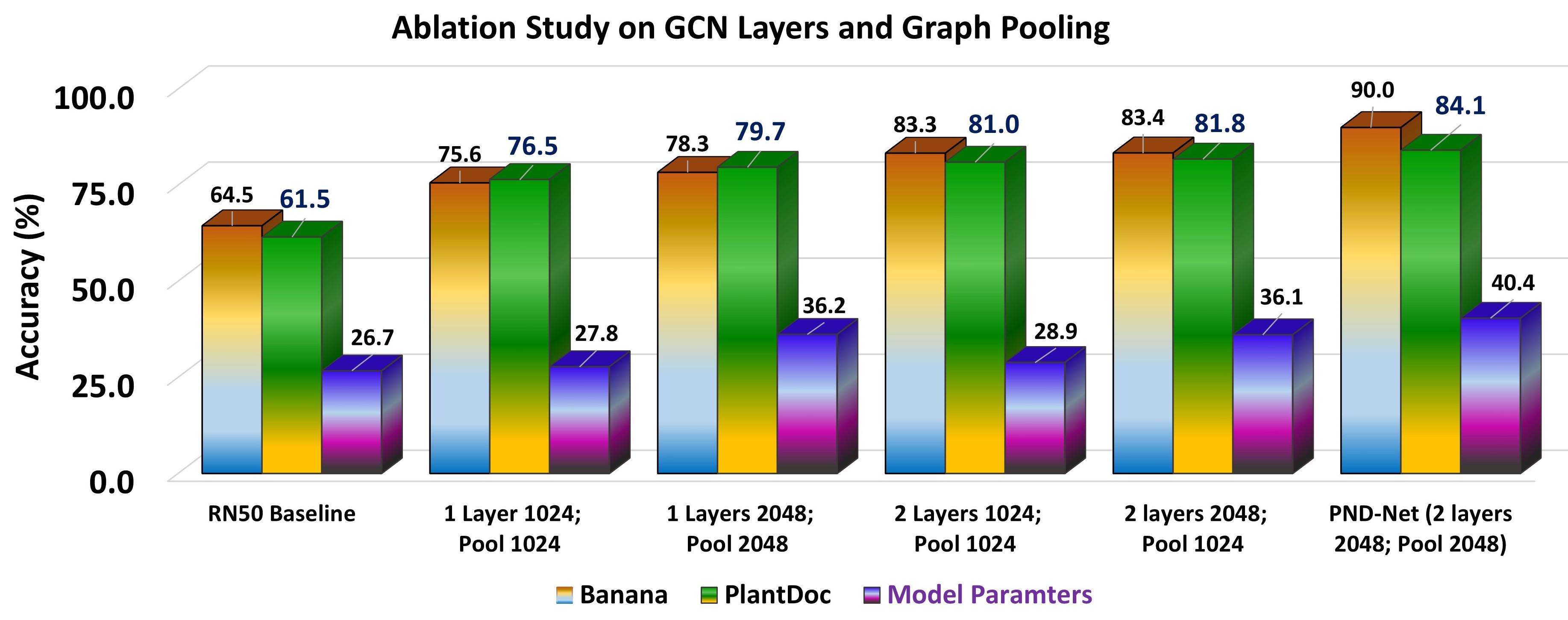}   }\hfill 
    }
\caption{a)  The performances of various formulations of the numbers of regions and spatial pyramid pooling feature vectors; b) The performances of different channel-wise node features within  GCN layers activation and propagation in the proposed method using the ResNet-50 backbone.    
}
\label{Acc_graph}
\end{figure}

It is evident that the GCN module indeed improves performance remarkably. In the case of the Banana dataset using Xception backbone, the  accuracy  of PND-Net is 89.25\%. Whereas,  averting  GCN layers, the degraded accuracy is 81.46\%, implying 7.79\% drop in accuracy. Even though, one GCN layer (Banana: 86.0\%) does not suffice to render the state-of-the-art performance on these plant datasets. The results of considering one layer GCN on all datasets are demonstrated in Table \ref{Abln1L}. Indeed, two layers in  GCN are beneficial in enhancing the performance over one  GCN layer, which is evident in the literature  \cite{bera2022sr}.  Hence, two GCN layers are included in the proposed PND-Net model architecture.

A comparative study on different number of regions and the number of pyramid pooled feature vectors using ResNet-50 is shown in Fig. \ref {Acc_graph}.(a), which clearly implies a gradual improvement in accuracy on the PlantDoc and Banana datasets. Lastly, the influences of different feature vector sizes in  GCN layer activations have been studied. In this study, the  channel dimensions of feature vectors 1024 and 2048 have been chosen for building the  graph structures using ResNet-50 backbone, implying the same channel dimensions have been considered in the PND-Net architecture. The results (Fig. \ref{Acc_graph}.(b) of such variations provide insightful implications about the performance of GCN layers.

The performances of  PND-Net with GCN output feature vector size of 1024 have been summarized in Table  \ref{Abln_GCN1024}. The results are very competitive with GCN's size of 2048. Thus, the model with 1024 GCN feature size could be preferred considering a trade off between the model parametric capacity with the performance.   
The detailed experimental studies imply overall performance boost on all datasets, and the proposed PND-Net achieves state-of-the-art results. In addition, new public datasets have been benchmarked for further enhancement. 

{However,  other categories of images such as high resolution, hyperspectral, etc.  have not been evaluated. One  reason is unavailability of such plant datasets for public research. Also,  data modalities such as soil-sensor information could be utilized for developing fusion based approaches.  Several existing ensemble methods have used  multiple backbones, which suffer from a higher computational complexity. Though, our method performs better than several existing works, yet, the computational complexity regarding model parameters and size of PND-Net could be improved. The reason is plugging the GCN module upon the backbone CNN, which incurs more parameters. To address this challenge, the graph convolutional layer could be simplified for reducing the model complexity. In addition, more realistic agricultural datasets representing field conditions such as occlusion, cluttered backgrounds, lighting variations, and others could be developed.  These  limitations of the proposed PND-Net will be explored in the near future. } 

\section{Conclusion}  \label{conclusion}
In this paper, a deep network called PND-Net has been proposed  for plant nutrition deficiency recognition using a GCN module, which is added on the top a CNN backbone. The performances have been evaluated  on four image datasets representing the plant nutrition deficiencies and leaf diseases. These datasets have recently been introduced publicly for assessment. The network has been generalized by building the deep network using four standard backbone CNNs, and the network architecture has been improved by incorporating pyramid pooling over region-pooled feature maps and feature propagation via a GCN.  We are the first to  evaluate these nutrition inadequacy datasets for monitoring plant health and growth. Our method has attained the state-of-the-art performance on the PlantDoc dataset for plant disease recognition.  We encourage the researcher  for further enhancement on these public datasets for  early stage detection of plant abnormalities, essential for sustainable agricultural growth. Furthermore, experiments have been conducted on the BreakHis (40X and 100X magnifications) and SIPaKMeD datasets, which are suitable for human health diagnosis. The proposed PND-Net have attained enhanced performances on these datasets too.  In the future, new deep learning methods would be developed for early stage disease detection of plants and health monitoring with balanced nutrition using other data modalities and imaging techniques.  

\vspace{ 1.0 cm}

\bibliography{sample.bib}
\vspace{ 1.0 cm}
\section*{Acknowledgements}

This work is supported by the New Faculty Seed Grant (NFSG) and Cross-Disciplinary Research Framework (CDRF: C1/23/168) projects, and Open Access facilities with necessary computational infrastructure at the Birla Institute of Technology and Science (BITS) Pilani, Pilani Campus, Rajasthan, 333031 India. This work has been also supported in part by the project (2024/2204), Grant Agency of Excellence, University of Hradec Kralove, Faculty of Informatics and Management, Czech Republic.

\vspace{ 0.5 cm}

\section*{Data Availability} 

The six datasets that support the findings which were used in this work are available using the given links. 

The Nutrient Deficient of Banana Plant dataset \cite{sunitha2023fully}  is collected from  \textit{https://data.mendeley.com/datasets/7vpdrbdkd4/1}.

The CoLeaf-DB dataset \cite{tuesta2023coleaf} for coffee leaf nutrition deficiency  classification is available at

\textit{https://data.mendeley.com/datasets/brfgw46wzb/1}. 

The Potato Leaf Disease Dataset \cite{shabrina2024novel} is available at \textit{https://data.mendeley.com/datasets/ptz377bwb8/1}. 

The PlantDoc 
 dataset \cite{singh2020plantdoc} is available at \textit{ https://github.com/pratikkayal/PlantDoc-Dataset}.  
 
 The BreakHis dataset \cite{spanhol2015dataset} is available at
 
 \textit{
https://web.inf.ufpr.br/vri/databases/breast-cancer-histopathological-database-breakhis/}, and 

can also be downloaded from 
\textit{ https://data.mendeley.com/datasets/jxwvdwhpc2/1}.

The original SIPaKMeD dataset  \cite{plissiti2018sipakmed}  can be found at \textit{ https://www.cs.uoi.gr/~marina/sipakmed.html}, and 

Kaggle \textit{https://www.kaggle.com/datasets/mohaliy2016/papsinglecell}.

%The corresponding author is responsible for submitting a \href{http://www.nature.com/srep/policies/index.html#competing}{competing interests statement} on behalf of all authors of the paper. This statement must be included in the submitted article file.

\section*{Competing interests}
The authors declare no competing interests.

\end{document}